\documentclass{article}

\usepackage{arxiv}

\usepackage[utf8]{inputenc} 
\usepackage[T1]{fontenc}  
\usepackage{hyperref}    
\usepackage{url}      
\usepackage{booktabs}    
\usepackage{amsfonts}    
\usepackage{nicefrac}    
\usepackage{microtype}   
\usepackage{lipsum}		
\usepackage{graphicx}
\usepackage{float}
\usepackage{natbib}
\usepackage{doi}
\usepackage{longtable}
\usepackage{xltabular} 
\usepackage{booktabs}
\usepackage{multirow}
\usepackage{amsmath}
\usepackage{graphicx}
\usepackage{makecell}
\usepackage{tcolorbox}
\tcbuselibrary{breakable}
\usepackage{amsmath}
\usepackage{amssymb}

\title{From Static Risk to Dynamic Disease Trajectories: Toward World-Model-Inspired Clinical Prediction}

\author{
Pujun Feng$^{1,9,\dagger}$, Xiaoyu Guo$^{2,9,\dagger}$, Seyed Ehsan Saffari$^{10,11,\ddagger}$, Min Hun Lee$^{3,\ddagger}$,\\
Siew-Kei Lam$^{4}$, Erik Cambria$^{4}$, Xibin Sun$^{5}$, Yangtao Zhou$^{6}$, Tong Yang$^{7}$,\\
Xiaoyu Zhang$^{8}$, Tao Tan$^{1}$, Yue Sun$^{1,\ddagger}$, Bin Cui$^{7,*}$\\
\small $^1$ Faculty of Applied Sciences, Macao Polytechnic University, Macao SAR 999078, China.\\
\small $^2$ School of Software \& Microelectronics, Peking University, Beijing 100871, China.\\
\small $^3$ School of Computing and Information Systems, Singapore Management University, Singapore 178902, Singapore.\\
\small $^4$ College of Computing and Data Science, Nanyang Technological University, Singapore 639798, Singapore.\\
\small $^5$ School of Public Health, Peking University, Beijing 100191, China.\\
\small $^6$ School of Computer Science and Technology, Xidian University, Xi’an 710126, China.\\
\small $^7$ School of Computer, Peking University, Beijing 100871, China.\\
\small $^8$ School of Computer Science and Technology, Beijing Institute of Technology, Beijing 100081, China.\\
\small $^9$ Medin.ai, Beijing 100871, China.\\
\small $^{10}$ Centre for Biomedical Data Science, Duke-NUS Medical School, National University of Singapore, Singapore 169857, Singapore.\\
\small $^{11}$ Duke-NUS AI + Medical Sciences Initiative, Duke-NUS Medical School, Singapore 169857, Singapore.\\
\small $^*$ Corresponding author: Bin Cui (\texttt{bin.cui@pku.edu.cn}).\\
\small $^\dagger$ Equal contribution. \quad $^\ddagger$ Co-corresponding author.\\
\small Emails: \texttt{fengpujun1@gmail.com}, \texttt{xiaoyu.guo@medin.ai}, \texttt{ehsan.saffari@duke-nus.edu.sg},\\
\small \texttt{mhlee@smu.edu.sg}, \texttt{ASSKLam@ntu.edu.sg}, \texttt{cambria@ntu.edu.sg}, \texttt{sunxibin2000@163.com},\\
\small \texttt{zhou\_yt@stu.xidian.edu.cn}, \texttt{yang.tong@pku.edu.cn}, \texttt{1120242616@bit.edu.cn},\\
\small \texttt{taotanjs@gmail.com}, \texttt{yuesun@mpu.edu.mo}, \texttt{bin.cui@pku.edu.cn}
}
\date{}

\renewcommand{\headeright}{}
\renewcommand{\undertitle}{}
\renewcommand{\shorttitle}{\textit{arXiv} Template}

\begin{document}
\maketitle

\begin{abstract}
  Clinical decision-making is a feedback system in which risk estimates influence treatment, treatment changes subsequent disease trajectories, and both shape what and when clinicians measure. Static prediction therefore often fails at the bedside: models trained on observational care logs can conflate disease biology with clinician behavior, especially under treatment confounder feedback and irregular or informative observation.

	This Review focuses on intervention-aware disease trajectory modeling in clinical AI, defined as methods that estimate patient-specific longitudinal disease evolution and assess how trajectories change under alternative treatment strategies. We organize the field around six linked components: three decision tasks (factual trajectory forecasting, counterfactual trajectory estimation, and policy evaluation) and three data-generating mechanisms (disease evolution, treatment assignment, and the observation process) that determine identifiability.

	We provide the first unified framework that bridges forecasting, counterfactual trajectories, and policy-level evaluation across discrete and continuous time with explicit treatment of treatment assignment, time-varying confounding, and observation-process bias. We synthesize multistate and joint models, temporal point-process and continuous-time methods, deep sequence architectures, and longitudinal causal inference; map each method family to the component it resolves; and align evaluation with claim strength using overlap diagnostics, uncertainty quantification, off-policy robustness, and target-trial-aligned validation. This synthesis shifts progress from benchmark prediction to decision-grade clinical evidence and enables capabilities that were previously difficult to achieve: treatment-sensitive individualized futures, pre-deployment stress-testing of clinical policies, and safer closed-loop learning health systems that can adapt and abstain when evidence is insufficient.
\end{abstract}

\keywords{Dynamic disease trajectories \and artificial intelligence \and risk of disease \and prediction model}



\section{Introduction}

\begin{figure}[t]
 \centering
 \includegraphics[width=0.95\linewidth]{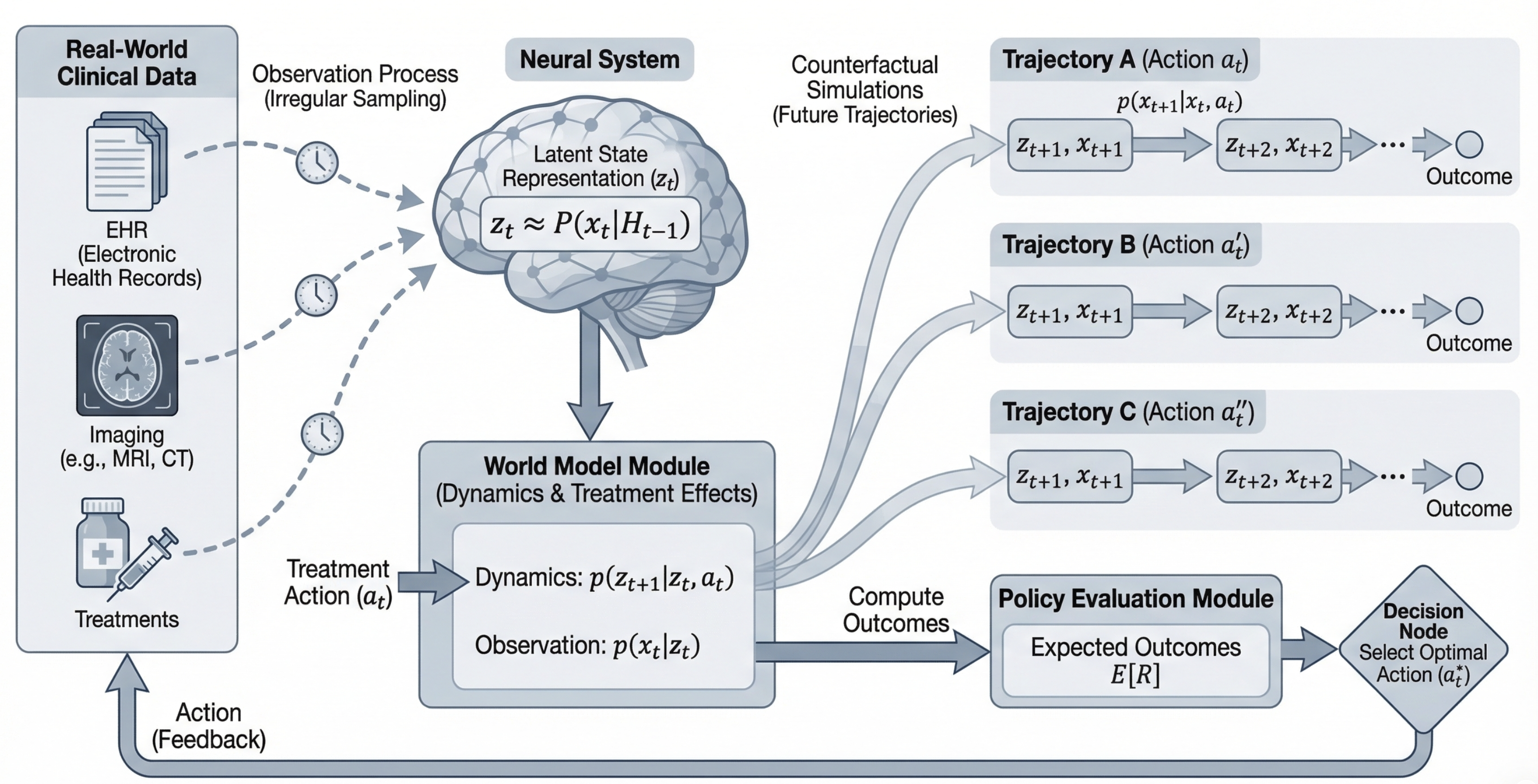}
 \caption{World-model-based prediction and decision-making in clinical trajectories.
Multimodal clinical data are encoded into a latent state \( z_t \) that summarizes patient history under an irregular observation process. A world model parameterizes disease dynamics via \( p(z_{t+1} \mid z_t, a_t) \) and observation generation via \( p(x_t \mid z_t) \). From the current state, the model performs counterfactual rollouts under alternative treatment actions, generating future trajectories and associated outcomes. These simulations support policy evaluation through expected outcomes \( \mathbb{E}[R] \), enabling selection of an optimal action \( a_t^* \). The resulting closed-loop system integrates prediction, intervention, and feedback, reframing trajectory modeling as an internal simulation problem for decision support.}
 \label{fig:intro_worldmodel}
\end{figure}

\begin{figure}[t]
 \centering
 \includegraphics[width=\linewidth]{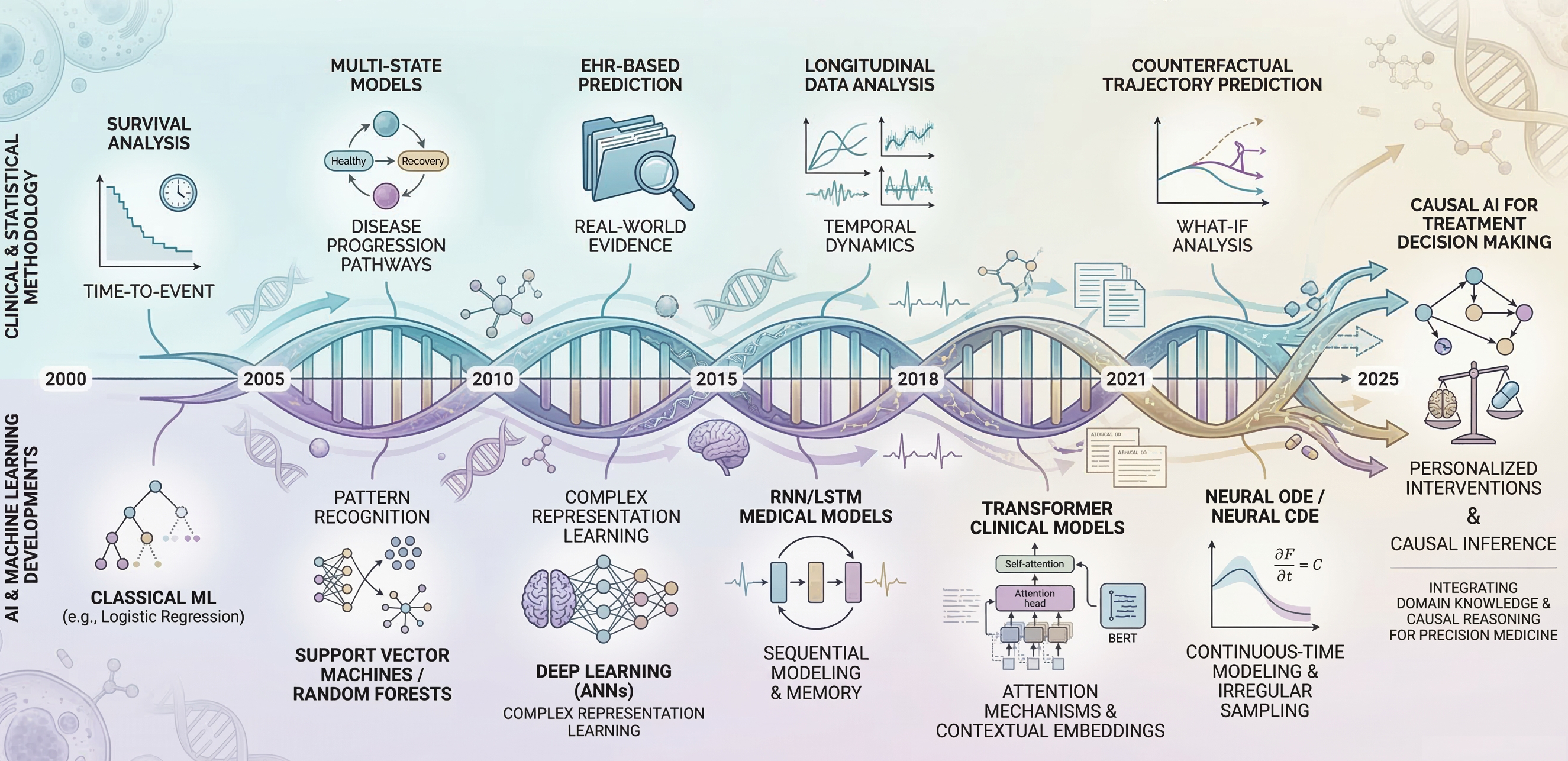}
 \caption{Evolution of disease trajectory modeling in medicine and AI.}
 \label{fig:intro_timeline}
\end{figure}

Identifying how diseases evolve over time, including their pathophysiological, clinical progression, and prognostic patterns, is central to modern medicine. Diseases, whether acute infections or chronic diseases such as diabetes, cancer, and cardiovascular disorders, rarely follow linear paths. Instead, they display dynamic temporal behaviors marked by relapses, remissions, and varied responses to treatment. Capturing and predicting these patterns, collectively known as disease trajectory analysis (DTA), is essential for precision medicine, early intervention, and population health management~\cite{intro_1}. Figure~\ref{fig:intro_timeline} provides a timeline of this evolution, from early feasibility studies in longitudinal prediction to current intervention-aware and closed-loop clinical AI paradigms.

Traditional approaches to characterizing disease progression have long depended on static clinical endpoints or snapshot biomarkers, which frequently fail to capture longitudinal variability. The rise of artificial intelligence (AI) and machine learning (ML) has revolutionized this field by enabling dynamic modeling of patient trajectories with high-dimensional, multimodal data streams that include electronic health records (EHRs), medical imaging, wearable sensor data, and omics datasets~\cite{trajvis2024jamia,intro_6}.

Recent high-quality studies have started to discuss world-model-like directions in medicine while still grounding claims in concrete clinical tasks rather than broad analogies. Foundation-model perspectives emphasize reusable longitudinal representations across heterogeneous medical tasks~\cite{moor2023gmai,guo2024sharedfm}, and newer medical world-model studies begin to simulate long-horizon patient evolution at population or individual levels~\cite{bica2025naturalhistory,yang2025medicalworldmodel,mu2026ehrworld}. Across these strands, dynamic prediction is the core capability, namely estimating how patient states evolve over time and how those trajectories change under different treatment contexts (Figure~\ref{fig:intro_worldmodel}).

Recent advances further demonstrate that AI-based longitudinal analysis, which follows the same patient across repeated observations over time rather than relying on a single cross-sectional snapshot, can accurately infer transitions between disease states, detect subtle shifts preceding clinical deterioration, and stratify patients according to personalized risk trajectories~\cite{linhas,wuuhar}. However, trajectory modeling becomes clinically actionable only when it is embedded in a complete closed-loop pathway: \textbf{Prediction $\rightarrow$ Decision $\rightarrow$ Execution $\rightarrow$ Feedback Learning}. In this loop, predictive models estimate near- and long-term risks, clinical decision modules map those predictions to candidate interventions, care teams execute selected actions in real workflows, and post-intervention outcomes are used to update models and policies in a learning health system~\cite{komorowski2018aiclinician,gottesman2019rlhc,wiens2019donoHarm,friedman2015lhs}. In parallel, visual analytics systems, such as TrajVis for chronic kidney disease, have enhanced interpretability by mapping complex model outputs into clinically meaningful visual forms~\cite{trajvis2024jamia}. Together, these developments move the field from one-off risk scoring toward continuously updated, intervention-aware learning systems.

Multimodal clinical data ecosystems are now routine, combining EHR histories, imaging repositories, wearable sensing streams, and other real-world evidence sources. This expansion creates genuine opportunity: richer longitudinal state representations can improve disease progression modeling, detect earlier inflection points, and support more personalized forecasting~\cite{intro_6}. Yet the same richness also creates a decision-level challenge. Once trajectory models inform treatment choice rather than retrospective risk scoring alone, heterogeneity across modalities, irregular sampling, missingness, and limited interoperability can directly weaken inference about action-consequence relationships~\cite{tripathi2025}. In this setting, fairness, interpretability, and regulatory compliance are not peripheral implementation concerns; they are core conditions for whether model outputs can be used responsibly in care pathways~\cite{badidi2023}.

This tension helps explain why existing reviews, while foundational, still leave an important synthesis gap. Early trajectory-focused reviews established the promise of AI for longitudinal patient modeling~\cite{intro_1}, and later methodological surveys clarified longitudinal EHR pipelines and common evaluation practice~\cite{carrascoRibelles2023jamia_review,intro_3}. However, most prior syntheses remain weighted toward forecasting under observed care, with less systematic treatment of intervention-aware decision support, counterfactual trajectory estimation, and closed-loop deployment under real clinical constraints~\cite{feuerriegel2024natmed,gottesman2019rlhc,wiens2019donoHarm}. At the same time, survey standards are becoming more rigorous, emphasizing explicit research questions, transparent selection logic, and traceable links between taxonomy and evidence~\cite{jin2025tsfm_survey}. We position this Review within that stricter standard.

To close this gap, we organize the field through a chronological technical-evolution lens (Figure~\ref{fig:intro_timeline}) that maps methodological progress to the strength of clinical claims each phase can support. Between 2016 and 2018, the dominant question was technical feasibility: could routine EHR histories be transformed into stable longitudinal representations for dynamic prediction? Early deep EHR systems answered this in the affirmative and established sequence learning as a practical backbone for patient-trajectory modeling~\cite{miotto2016deeppatient,choi2016doctorai,choi2016retain,rajkomar2018}. The main scientific output of this phase was predictive capacity under observed care.

From 2019 to 2021, attention shifted from representation to decision relevance. Sepsis policy-learning studies and accompanying methodological critiques showed that strong predictive discrimination does not automatically imply safe treatment recommendations~\cite{komorowski2018aiclinician,raghu2017sepsis_rl,gottesman2019rlhc}. In parallel, transformer pretraining improved transfer across longitudinal tasks but amplified concerns about calibration, transportability, and interpretability in real workflows~\cite{li2020behrt,rasmy2021}. Since 2022, the frontier has moved further toward intervention-aware trajectory science: continuous-time formulations, informative-observation correction, and target-trial-aligned validation now frame progress as \emph{decision-grade evidence generation} rather than architecture-level benchmarking~\cite{pmlr-v162-seedat22b,pmlr-v202-vanderschueren23a,hernan2016targettrial,feuerriegel2024natmed}. Recent population-scale generative transformer evidence further strengthens this shift, showing that pretraining on longitudinal disease histories can capture natural disease progression patterns and improve long-horizon incidence forecasting across external cohorts~\cite{bica2025naturalhistory}.

Accordingly, this Review is designed to contribute beyond prior summaries in three ways. First, we reframe disease trajectory modeling as a full clinical control loop, namely Prediction $\rightarrow$ Decision $\rightarrow$ Execution $\rightarrow$ Feedback Learning, rather than a sequence prediction task alone. Second, we unify methodological strands that are usually discussed separately (multistate/joint models, deep temporal sequence models, and longitudinal causal inference for dynamic treatment regimes), with explicit attention to feedback between treatment and confounders, irregular observation, and transportability assumptions~\cite{murphy2003dtr,robins2000msm,feuerriegel2024natmed}. Third, we emphasize decision grade evaluation, including target-trial-aligned validation, off-policy considerations, uncertainty-aware abstention, and overlap diagnostics, so that claims are aligned with intended clinical use~\cite{hernan2016targettrial,gottesman2019rlhc,pmlr-v151-de-brouwer22a}.

To formalize these contributions, we organize the Review around four tightly linked research questions that form a dependency chain from scientific target definition to translational evidence:

\begin{enumerate}
  \setlength{\itemsep}{0.25em}
  \setlength{\parsep}{0pt}
  \setlength{\parskip}{0pt}
  \setlength{\topsep}{0.3em}
  \item \textbf{RQ1: What is the intervention-aware scientific object of disease trajectory modeling in real clinical care?}
  We ask how trajectories should be defined as joint longitudinal objects over latent disease evolution, treatment decisions, and observation processes, and which estimands (factual trajectories, counterfactual trajectories under dynamic regimes, and policy value) are clinically meaningful.

  \item \textbf{RQ2: Under which causal and data-support assumptions are these estimands identifiable in real-world longitudinal data?}
  We examine sequential exchangeability, consistency, and positivity under treatment--confounder feedback, irregular and informative observation, censoring, and transport shift, and we specify diagnostics that separate weak empirical support from model misspecification.

  \item \textbf{RQ3: Which modeling and estimation strategies preserve this causal semantics across discrete and continuous time?}
  We compare g-method-based estimators, representation-balancing sequence models, continuous-time/event-process formulations, and uncertainty-aware hybrids in terms of what they identify, where they extrapolate, and how multimodal irregularity changes bias--variance and interpretability trade-offs.

  \item \textbf{RQ4: What constitutes decision-grade evidence for safe translation into learning health systems?}
  We define a staged evidence program that links estimands to validation design (target-trial emulation, quasi-experimental triangulation, and off-policy evaluation), quantifies uncertainty and overlap, and extends to post-deployment monitoring, abstention, and safety governance.
\end{enumerate}

\begin{figure}[!htbp]
 \centering
 \includegraphics[width=0.95\linewidth]{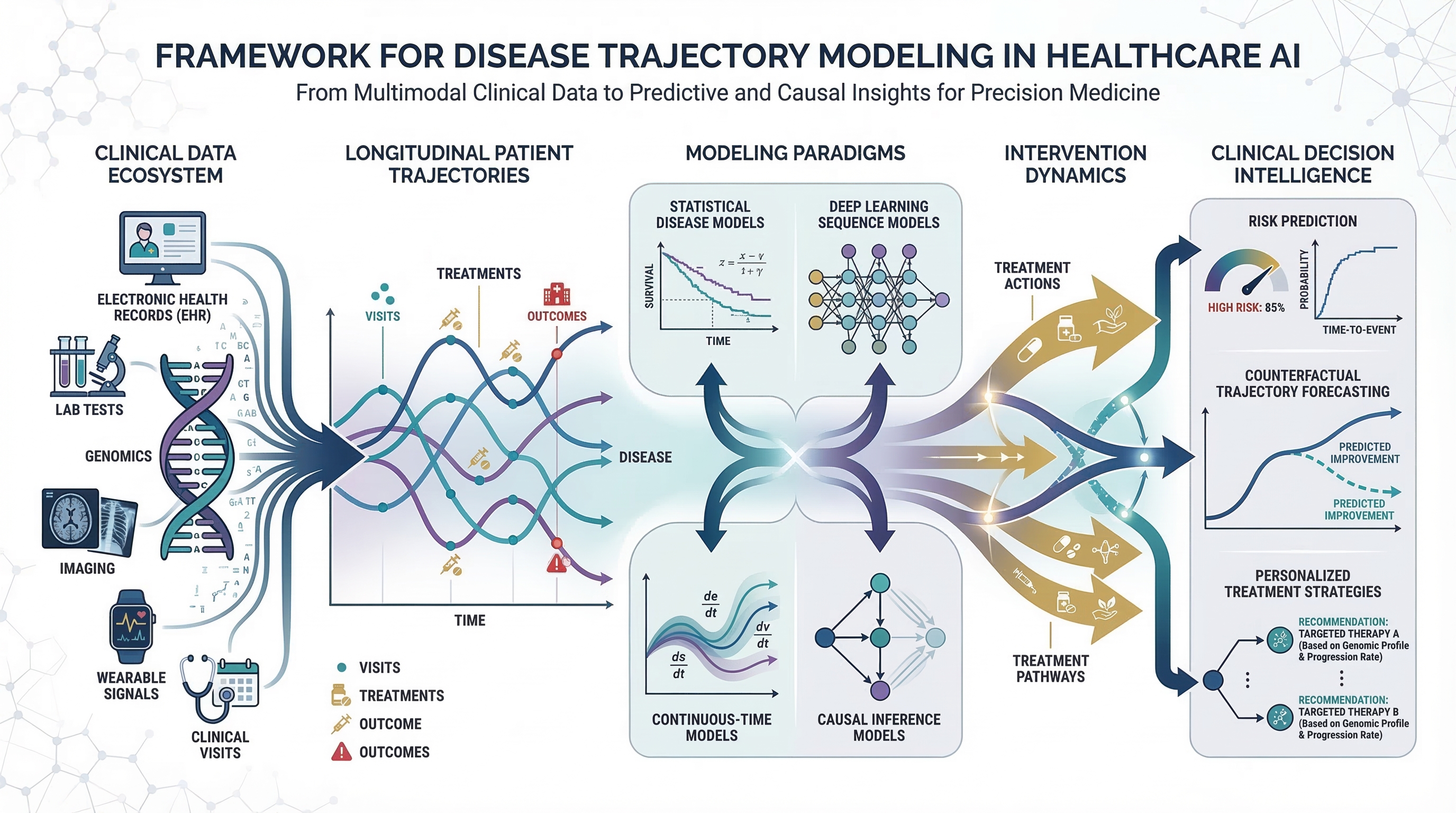}
 \caption{Overview of our intervention-aware disease trajectory modeling framework.}
 \label{fig:overview_framework}
\end{figure}

By integrating insights across biomedical informatics, data science, and
translational medicine, we map a path toward AI-enabled longitudinal health intelligence that anticipates rather than merely describes disease evolution. Figure~\ref{fig:overview_framework} should be interpreted as this end-to-end closed loop: multimodal longitudinal data support trajectory prediction; predictions inform treatment decisions; decisions are translated into clinical execution; and real-world outcomes feed back into model updating, policy refinement, and continuous system learning~\cite{gottesman2019rlhc,wiens2019donoHarm,friedman2015lhs}.

\section{Literature Search and Study Selection}

A systematic literature search was conducted to identify studies on artificial intelligence–based disease trajectory modeling and clinical risk prediction. The search was performed across three major databases: \textit{Web of Science}, \textit{Scopus}, and \textit{PubMed}. The following query terms were used:

\begin{quote}
("large model" OR "foundation model" OR "artificial intelligence" OR "transformer" OR "Large Language Model" OR "LLM*" OR "AI-based" OR "Representation model*" OR "predictive model*") \\
AND \\
("disease prognosis" OR "patient trajectory" OR "disease progression" OR "disease risk" OR "clinical risk" OR "medical risk" OR "disease diagnosis" OR "disease detection" OR "mortality risk prediction" OR "predictive clinical" OR "electronic health record*")
\end{quote}

The search yielded a total of 1,241 records, including 109 from Web of Science, 831 from Scopus, and 301 from PubMed. After removing 360 duplicate records across databases, 881 unique studies remained for further screening.

Title and abstract screening were then conducted to assess relevance. The screening process was independently performed by two reviewers, one with a computer science background and one with a medical background. In cases of disagreement, a third reviewer with expertise in medical artificial intelligence was consulted to reach a consensus.

Studies unrelated to human diseases, including those focusing on animal or plant diseases, were excluded. After this step, 722 studies remained. Subsequently, studies focusing exclusively on medical image prediction without integrating clinical text or electronic health record data were removed, leaving 310 studies for full-text evaluation.

During the eligibility assessment, 191 studies were excluded for the following reasons: (1) not focused on disease prediction, diagnosis, or clinical intervention ($n=73$); (2) studies addressing non-human diseases ($n=25$); and (3) lack of clinical textual information or minimal use of electronic health record data ($n=93$).

\begin{figure}[!htbp]
\centering
\includegraphics[width=0.7\linewidth]{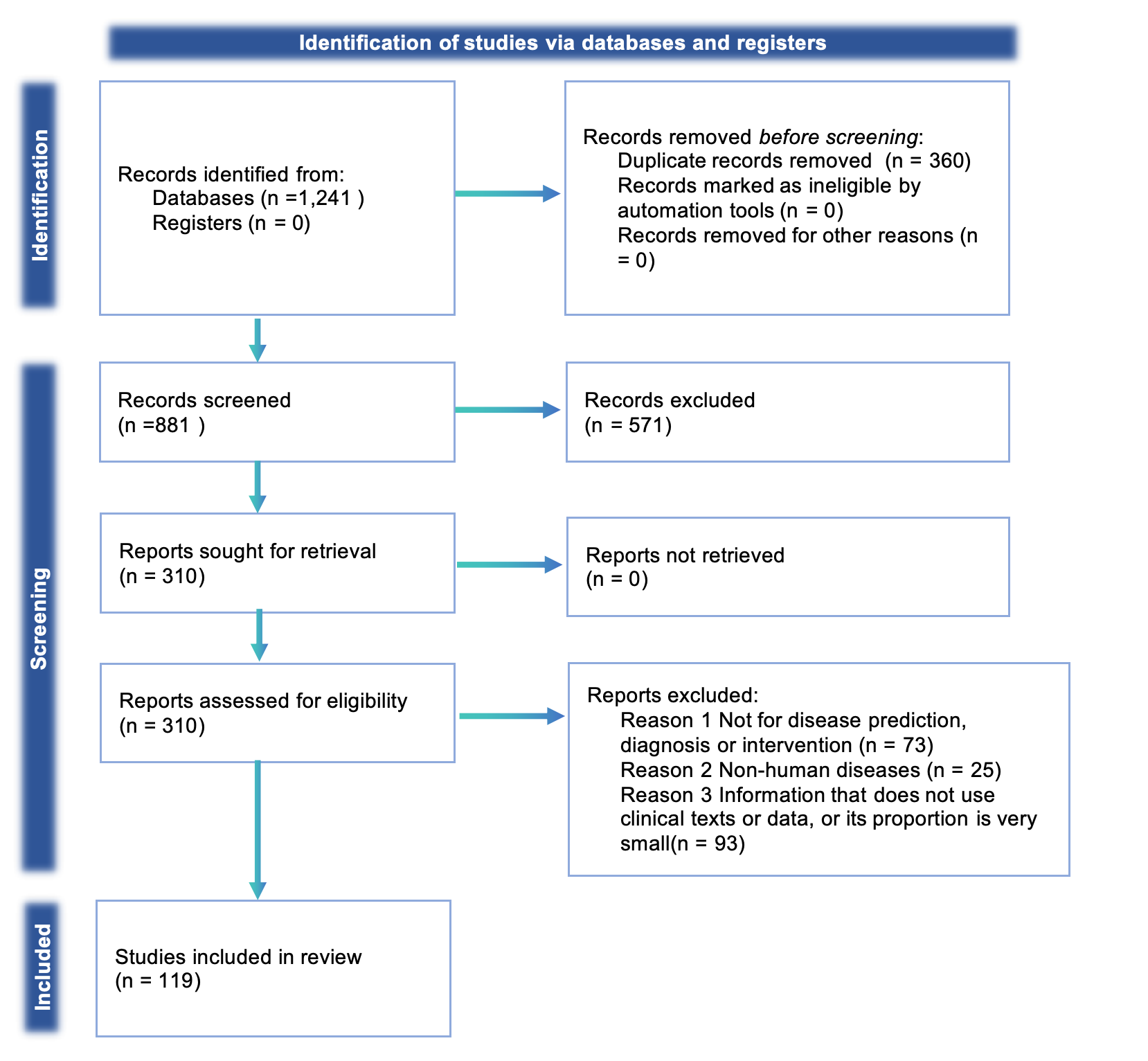}
\caption{PRISMA diagram for the screening process.}
\label{fig:screening}
\end{figure}

After applying all inclusion and exclusion criteria, 119 studies were retained for qualitative analysis in this review. The detailed study selection process is illustrated in Figure~\ref{fig:screening}.

\section{Conceptual and Review Landscape: Problem Framing and Taxonomy}
\label{sec:conceptual-review-landscape}

\begin{figure}[t]
 \centering
 \includegraphics[width=\linewidth]{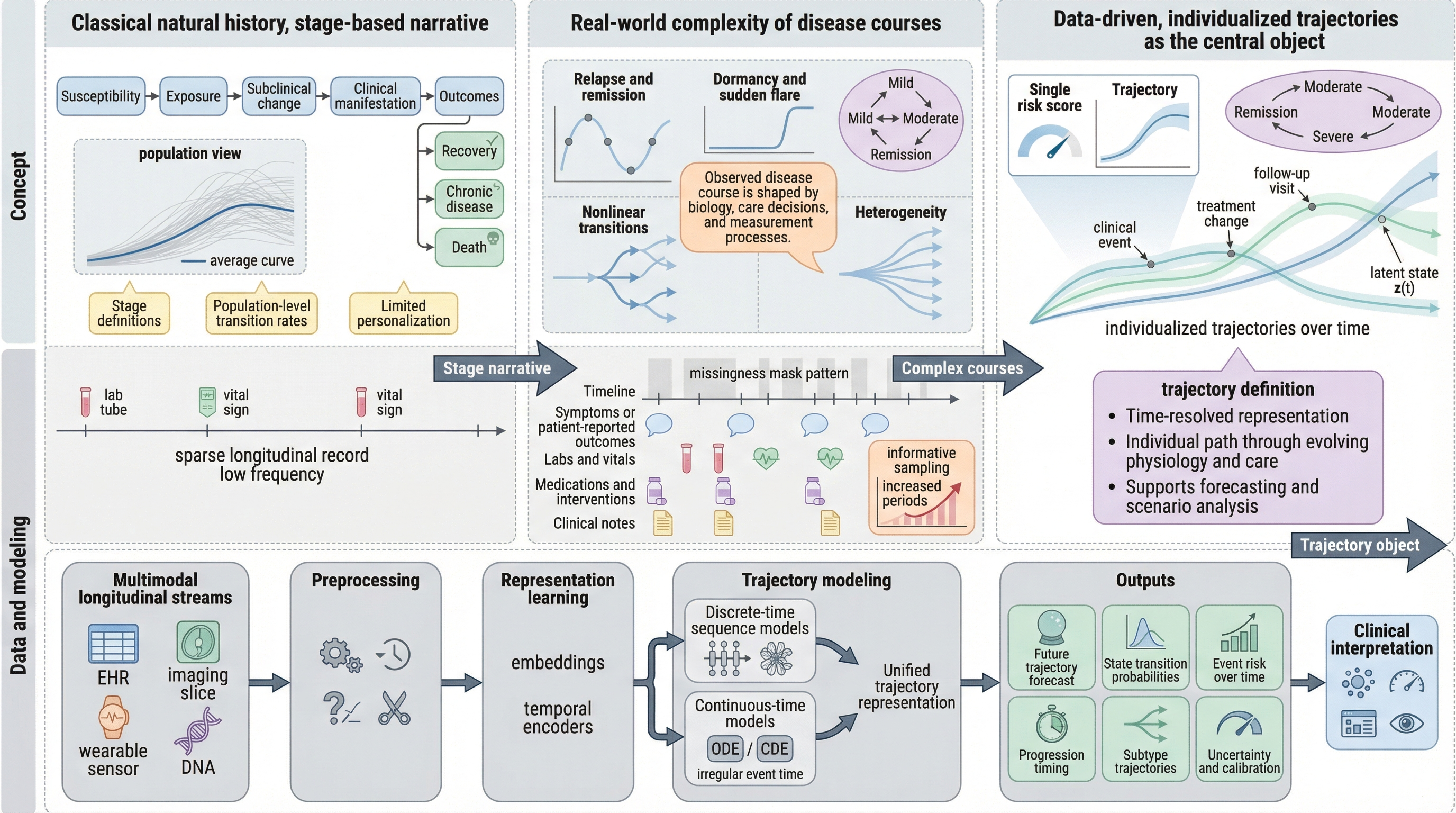}
 \caption{Overview: from the natural history of disease to data-driven trajectories.}
 \label{fig:2}
\end{figure}
Clinical thinking has long organized disease progression through the lens of natural history, using stage-based narratives to describe how populations move from susceptibility and exposure to subclinical changes, clinical manifestation, and downstream outcomes. However, real-world courses frequently deviate from a monotonic stage sequence. Many conditions exhibit relapse and remission, periods of apparent dormancy followed by abrupt deterioration, nonlinear transitions across severity states, and substantial heterogeneity across individuals. These complexities become even more pronounced once care is taken into account, since treatment adjustments and follow-up intensity are themselves responsive to evolving disease status. In this setting, the central object shifts from a single risk score or an average stage transition pattern to an individualized, time-resolved trajectory constructed from multimodal longitudinal data. Figure~\ref{fig:2} summarizes this conceptual transition and highlights how richer data and modern temporal modeling enable trajectory representations that better reflect the coupled dynamics of disease evolution, clinical actions, and observation processes.

To ground the downstream methodological discussion (Sections~\ref{3.1}-\ref{3.5}) without changing scope, we use a single integrated clinical example across two time scales. In ICU sepsis care, clinicians repeatedly titrate fluids, vasopressors, and antimicrobials over hours based on rapidly updated physiology; this creates explicit treatment--confounder feedback and adaptive policy updates over time~\cite{evans2021sepsis,komorowski2018aiclinician,gottesman2019rlhc}. In longitudinal type 2 diabetes care, treatment intensification evolves over years in response to HbA1c trajectories and comorbidity context, and cumulative glycaemic exposure is strongly associated with later complications~\cite{ukpds1998,ukpds2000hba1c}. Despite different tempos, both settings instantiate the same causal trajectory problem: identify counterfactual paths under dynamic interventions, diagnose overlap and transport limits, and evaluate decision utility rather than only one-step predictive discrimination~\cite{robins2000msm,2-17,hernan2016targettrial}. 

\subsection{From the natural history of disease to data-driven trajectories}
\label{3.1}
The classical natural history of disease describes progression along a broad continuum (susceptibility $\rightarrow$ exposure $\rightarrow$ subclinical change $\rightarrow$ clinical manifestation $\rightarrow$ resolution or chronicity), and historically grounded much of disease prediction in population-level stage transitions~\cite{porche2022}. 
This framing is foundational but often presumes comparatively stable stage structure and limited individual heterogeneity, which can be violated in modern complex diseases with relapse--remission patterns, dormancy, and nonlinear transitions~\cite{sypsa2021,costa2023,fabrizio2025}. 
As longitudinal biomedical data have become more abundant, the conceptual object of interest has shifted from a single risk score to a trajectory: an individual, time-resolved path through evolving physiology, clinical events, and care processes~\cite{intro_1,intro_3}. 
Trajectory-centric reviews argue that modeling should explicitly represent (i) the temporal evolution of patient state and (ii) the clinical actions that shape this evolution, rather than treating time as a nuisance covariate appended to a static prediction task~\cite{intro_1}. 

\subsection{The digital phenome: representation learning as a trajectory substrate}
\label{3.2}
The digital phenome includes multimodal, time-stamped information from structured EHR fields, clinical notes, and continuous physiologic streams. It provides the substrate on which modern trajectory models are trained~\cite{rajkomar2018,rasmy2021,kline2022}. 
Sequence encoders (e.g., RNNs, temporal convolutions, and transformers) map heterogeneous observations into latent histories that support forecasting and stratification over time~\cite{graves2012,cho2014,bai2018,rasmy2021}. 
This direction is exemplified by a recent Nature study that trained a generative transformer on large-scale longitudinal health records to model the natural history of human disease and support broad trajectory-level prediction tasks~\cite{bica2025naturalhistory}. 
However, major reviews emphasize that many "longitudinal" studies still evaluate at fixed horizons using conventional supervised metrics, which can conflate improved feature aggregation with genuine modeling of disease dynamics as a coherent process~\cite{kline2022,wuuhar}. 
Accordingly, conceptual work distinguishes \emph{trajectory-aware prediction} (learning time-varying risk or outcomes) from \emph{trajectory modeling} (learning an explicit temporal object whose evolution and uncertainty can be queried across time)~\cite{intro_1,intro_3}. 

Before formalizing the causal trajectory estimation problem, we provide a self-contained primer on the key concepts from longitudinal causal inference that underpin intervention-aware modeling (Box 1). Readers familiar with g-methods and sequential exchangeability may skip directly to Section~\ref{section:Causal}.

\begin{tcolorbox}[
title=BOX 1: Key Concepts in Longitudinal Causal Inference,
breakable,
colback=gray!5,
colframe=black,
fonttitle=\bfseries]

\textbf{Longitudinal causal inference} extends classical causal reasoning to settings where treatments, confounders, and outcomes evolve over time in a feedback loop. Understanding counterfactual trajectories—what would have happened under alternative treatment sequences—requires careful handling of this temporal dependence. The following concepts underpin intervention-aware disease trajectory modeling.

\bigskip

\textbf{1. Sequential Exchangeability (Time-Varying Unconfoundedness)}

\textbf{Definition.} At each time point $t$, treatment assignment $A_t$ is independent of potential future outcomes conditional on observed history $H_t$:

\[
A_t \perp Y^{\bar a}_{t+1:T} \mid H_t
\]

This assumption states that once all measured patient history (covariates, past treatments, past outcomes) is accounted for, the choice of treatment at time $t$ is “as good as random" with respect to unmeasured factors affecting future outcomes.

\textbf{Holds when.} Treatment decisions are based entirely on recorded clinical information (labs, vitals, prior treatment responses) in well-documented EHR systems.

\textbf{Fails when.} Clinicians act on unrecorded information (e.g., informal frailty assessments, family preferences not charted, tacit severity judgements), leading to hidden confounding and biased counterfactual estimates.

\textbf{Practical implication.} Models must condition on sufficiently rich patient history.

\bigskip

\textbf{2. Positivity / Overlap (Experimental Treatment Assignment)}

\textbf{Definition.}

\[
P(A_t=a \mid H_t) > 0
\]

For each observed history $H_t$, every treatment option must have positive probability.

\textbf{Violations in practice}

\begin{itemize}
\item Deterministic protocols (e.g., creatinine $>3.0$ always triggers dialysis)
\item Sparse treatment histories in long sequences
\item Severity-driven treatment selection
\end{itemize}

\textbf{Practical implication.} Overlap diagnostics (propensity distributions, effective sample size) should always be reported.

\bigskip

\textbf{3. Treatment--Confounder Feedback (Time-Varying Confounding)}

A variable $L_t$ that

\begin{itemize}
\item is affected by past treatment $A_{t-1}$
\item affects future treatment $A_t$
\item affects future outcomes $Y_{t+1}$
\end{itemize}

creates a temporal feedback loop.

\textbf{Example.}

At $t=0$ a patient receives a fluid bolus ($A_0$).  
At $t=1$ blood pressure improves ($L_1$).  
This influences future vasopressor treatment ($A_1$).

Here $L_1$ is both

\begin{itemize}
\item a mediator
\item a confounder
\end{itemize}

Standard regression adjustment cannot correctly handle this structure.

\textbf{Practical implication.} Specialized causal methods (g-methods, weighting, structural causal models) are required.

\bigskip

\textbf{4. The G-Formula (Standardization)}

The $g$-formula estimates counterfactual outcomes under treatment sequence $\bar a$:

\[
E[Y^{\bar a}] =
\sum_{\bar l}
\prod_t
P(L_t \mid H_{t-1},A_{t-1})
P(Y_t \mid H_t,A_t=\bar a_t)
\]

\textbf{Procedure}

\begin{enumerate}
\item Fit covariate evolution models
\[
P(L_t \mid H_{t-1},A_{t-1})
\]

\item Fit outcome models
\[
P(Y_t \mid H_t,A_t)
\]

\item Simulate trajectories while fixing $A_t=\bar a_t$
\end{enumerate}

Modern implementations often rely on neural sequence models (RNNs or Transformers).

\bigskip

\textbf{5. Marginal Structural Models with Inverse Probability Weighting}

\textbf{Steps}

\[
P(A_t \mid H_t)
\]

\[
w_t=\prod_{s=0}^{t}\frac{1}{P(A_s \mid H_s)}
\]

Then estimate a marginal outcome model using the weighted data.

Observations receiving unlikely treatments receive larger weights, approximating randomized assignment.

\bigskip

\textbf{6. G-Formula vs MSM}

\begin{center}
\begin{tabularx}{\textwidth}{p{3.8cm}X X}
\toprule
 & \textbf{G-Formula} & \textbf{MSM with IPW} \\
\midrule

\textbf{Mechanism}
& Forward simulation
& Inverse weighting \\

\textbf{What you model}
& Full data-generating process (confounders + outcomes)
& Treatment assignment + marginal outcome model \\

\textbf{Strengths}
& Natural for complex time-varying confounders; handles continuous treatments easily
& More robust to outcome model misspecification; transparent marginal effects \\

\textbf{Weaknesses}
& Requires correct specification of all intermediate distributions; compounding errors in long rollouts
& High-variance weights with poor overlap; requires correct propensity models at every time point \\

\textbf{When to prefer}
& Complex confounder evolution + good overlap
& Uncertain outcome model + well-understood treatment assignment \\

\textbf{Modern hybrid}
& \multicolumn{2}{X}{
Use deep networks for both propensity and outcome models, combined with stabilized weighting or doubly-robust estimation. In high-dimensional EHR settings, the distinction blurs—methods such as RMSN and Causal Transformer implicitly do both.
} \\

\bottomrule
\end{tabularx}
\end{center}

\bigskip

\bigskip

\textbf{7. Why These Concepts Matter for Trajectory Modeling}

Treating time-varying treatments as “just another feature” in $X_t$ fails under treatment–confounder feedback because (i) bias at time t propagates through future confounders and treatments (confounding accumulation); (ii) conditioning on post-treatment variables distorts causal effects (mediation blocking); and (iii) long treatment sequences erode overlap, making counterfactual queries unreliable (positivity erosion).

\textbf{Required guardrails.}

Explicit propensity modeling or balancing objectives; avoidance of post-treatment conditioning; routine reporting of overlap and positivity diagnostics for counterfactual queries; and validation against causal estimands rather than predictive accuracy on observed trajectories alone.

\textbf{Warning.}

Without these guardrails, trajectory models can achieve high predictive performance on logged data while producing systematically biased treatment-effect estimates—a particularly dangerous combination for clinical decision support.

\bigskip

\textbf{Further Reading}

\begin{itemize}

\item \textbf{Foundational.}  
Robins, J. (1986).  
"A new approach to causal inference in mortality studies with a sustained exposure period." 
\textit{Mathematical Modelling}, 7: 1393–1512.

\item \textbf{Tutorial.}  
Hernán, M. A., \& Robins, J. (2020).  
\textit{Causal Inference: What If}. Chapters 19–21.  
Available at:  
\url{https://www.hsph.harvard.edu/miguel-hernan/causal-inference-book/}

\item \textbf{ML integration.}  
Schulam, P., \& Saria, S. (2017).  
"Reliable decision support using counterfactual models."
\textit{Advances in Neural Information Processing Systems (NeurIPS)}.

\item \textbf{Continuous-time extension.}  
Xu, Y., et al. (2024).  
"Causal inference with functional longitudinal data."
\textit{Biometrika}.

\end{itemize}
\end{tcolorbox}

\subsection{Causal problem formulation: 
intervention-aware trajectories as potential-outcome paths}
\label{section:Causal}
For intervention-aware disease trajectory modeling, the dominant framing is longitudinal causal inference under dynamic treatment regimes, where the target is not merely forecasting the observed future but estimating counterfactual trajectories under alternative intervention sequences or policies~\cite{2-1,robins2000msm,2-17,murphy2003dtr}. 
In discrete time, for individual $i$ at time $t\in\{0,\dots,T\}$, let $X_{i,t}$ denote observed covariates, $A_{i,t}$ denote treatment/action, and $Y_{i,t}$ denote an outcome (possibly multivariate), with history $H_{i,t}=(X_{i,0:t},A_{i,0:t-1},Y_{i,0:t})$~\cite{2-1,robins2000msm}. 
A (possibly stochastic) dynamic regime is a sequence of decision rules $g=\{g_t\}_{t=0}^{T}$ mapping histories to actions, and the \emph{potential outcome trajectory} $Y_{i,0:T}^{g}$ denotes the path that would be realized under regime $g$~\cite{2-17,robins2000msm}. 
Identification under sequential consistency, sequential exchangeability given history, and positivity yields the longitudinal $g$-formula as a conceptual backbone for trajectory-level causal targets~\cite{robins2000msm,2-1}. 
One canonical discrete-time expression for a fixed treatment sequence $\bar a_{0:T}$ is
\begin{align}
\mathbb{E}\!\left[Y_T^{\bar a_{0:T}}\right]
&=
\int
\mathbb{E}\!\left[Y_T \mid \bar X_T=\bar x_T,\bar A_T=\bar a_T\right]
\prod_{t=0}^{T} p\!\left(x_t \mid \bar x_{t-1},\bar a_{t-1}\right)\,
d\bar x_T,
\label{eq:g-formula-discrete}
\end{align}
which makes explicit that causal trajectory estimation requires modeling covariate evolution under treatment--confounder feedback~\cite{robins2000msm,2-1}. 
Marginal structural models (MSMs) provide an alternative identification-and-estimation route via inverse-propensity (stabilized) weights, for example
\begin{align}
w_i
&=
\prod_{t=0}^{T}
\frac{p(A_{i,t}=a_{i,t}\mid \bar A_{i,0:t-1}=\bar a_{i,0:t-1})}
   {p(A_{i,t}=a_{i,t}\mid \bar A_{i,0:t-1}=\bar a_{i,0:t-1},\,\bar X_{i,0:t}=\bar x_{i,0:t})},
\label{eq:iptw}
\end{align}
which reweights observed trajectories to emulate a randomized longitudinal experiment~\cite{robins2000msm,2-1}. 
These identification ideas have been operationalized in modern deep longitudinal models, for example recurrent marginal structural networks (RMSNs) that combine sequence modeling with MSM-style weighting in discrete time~\cite{2-11,robins2000msm}. 
Transformer-based architectures further extend multi-step counterfactual prediction by modeling long-range temporal dependencies while retaining an explicit potential-outcome target, as in the Causal Transformer framework~\cite{2-3}. 
Continuous-time trajectory modeling is motivated by irregular clinical sampling and event-driven interventions, for which neural controlled differential equation models and continuous-time weighting ideas have emerged as practically important instantiations of the same causal targets~\cite{pmlr-v162-seedat22b,pmlr-v202-vanderschueren23a}. 
Recent causal theory for functional longitudinal data formalizes dynamic treatment effects as estimands over high-frequency trajectories, clarifying identifiability and providing principled targets for continuous-time intervention-aware models~\cite{pmlr-v236-ying24a}. 
High-impact clinical perspectives further emphasize that decision support requires causal estimands, explicit assumptions, and careful assessment of overlap and generalizability, as purely predictive accuracy can be misleading for treatment selection~\cite{feuerriegel2024natmed,2-17}. 

\subsection{A taxonomy for the review landscape}
Trajectory-centric reviews explicitly unify disparate modeling paradigms, including state-space models, temporal point processes, deep sequence encoders, and causal/policy frameworks, under a patient trajectory lens~\cite{intro_1}.

Complementary scoping work emphasizes that trajectory analysis spans both methodology (how to represent and estimate trajectories) and translation (how to evaluate and visualize trajectories for decision support)~\cite{intro_3,trajvis2024jamia}. 
To make the landscape comparable across communities, a useful first-order taxonomy separates four strata: (S1) conceptual reviews/frameworks, (S2) trajectory models without explicit counterfactual semantics, (S3) intervention-aware causal models in discrete time, and (S4) intervention-aware causal models in continuous time~\cite{intro_1,2-1,pmlr-v162-seedat22b}. 

This stratification distinguishes whether interventions are treated as covariates versus manipulable objects defining counterfactual worlds, and whether time is represented as visit-indexed sequences versus continuous-time processes~\cite{intro_1,2-1}. Beyond strata, expert comparison benefits from explicit axes that declare what trajectory and what intervention a model supports~\cite{intro_1,2-1}.
Table~\ref{tab:taxonomy-axes} summarizes a compact set of axes that recur across high-level reviews and representative methodological lines~\cite{intro_1,2-1}. 

\begin{table}[t]
\caption{Practical taxonomy axes for dynamic, intervention-aware disease trajectory modeling}
\centering
\small
\begin{tabular}{p{0.22\linewidth} p{0.52\linewidth} p{0.20\linewidth}}
\hline
\textbf{Approach} & \textbf{Key options (examples)} & \textbf{Representative refs.} \\
\hline
Time representation &
Discrete visits; continuous time with irregular timestamps; event-driven time & ~\cite{2-3,pmlr-v162-seedat22b,schulam2017reliable} \\
Trajectory target &
Horizon risk; full outcome path; latent state; multi-event paths & ~\cite{intro_1,intro_3} \\
Intervention semantics &
Treatments as covariates; fixed sequences; dynamic regimes/policies & ~\cite{robins2000msm,2-17,2-1} \\
Confounding regime &
Observed time-varying confounding; hidden confounding; sensitivity/bounds & ~\cite{robins2000msm,2-1,2-17} \\
Observation process &
Regular sampling; irregular missingness; informative visits/measurement intensity & ~\cite{intro_1,pmlr-v202-vanderschueren23a} \\
Uncertainty treatment &
Point forecasts; epistemic uncertainty for overlap/decision support; Bayesian posteriors & ~\cite{pmlr-v202-vanderschueren23a,pmlr-v162-seedat22b} \\
Evaluation focus &
Factual forecast error; counterfactual trajectory error; policy value/utility & ~\cite{2-1,intro_3,pmlr-v202-vanderschueren23a} \\
\hline
\end{tabular}

\label{tab:taxonomy-axes}
\end{table}

\subsection{Bottlenecks and methodologies}
\label{3.5}
\textbf{(i) Longitudinal prediction is not yet counterfactual trajectory modeling.}Trajectory reviews stress that merely feeding repeated measurements into a predictor does not yield an explicit trajectory object, nor does it support counterfactual queries under interventions~\cite{intro_1,intro_3}.Intervention-aware work therefore makes the potential-outcome trajectory (or downstream policy utility) the explicit estimand, rather than treating treatment as a standard feature~\cite{2-1,robins2000msm}.

\textbf{(ii) Time-varying confounding is structural, not incidental.} 
When confounders both affect and are affected by past treatment, naive adjustment can induce bias and motivates $g$-methods (e.g., MSM/IPW, $g$-computation) and their modern ML instantiations~\cite{robins2000msm,2-1}. 
Accordingly, reviews stress aligning model objectives with the causal estimand, checking positivity/overlap, and using evaluation protocols that reflect the intended counterfactual query (trajectory-level or policy-level), rather than relying solely on factual forecasting metrics~\cite{2-1,2-17,feuerriegel2024natmed}. 

\textbf{(iii) Irregular sampling and informative observation processes.} 
EHR trajectories are irregular and partially observed, so continuous-time formulations and observation-process correction can be essential for unbiased counterfactual forecasting over time~\cite{intro_1,pmlr-v162-seedat22b,pmlr-v202-vanderschueren23a}. 
In this view, visit times are not merely missingness but part of the data-generating process, and inverse-intensity weighting provides one principled route to adjust for informative sampling in continuous-time trajectory models~\cite{pmlr-v202-vanderschueren23a}. 

\textbf{(iv) Translational bottlenecks: evaluation, interpretability, and decision support.} 
Scoping reviews highlight that trajectory methods are evaluated using heterogeneous metrics and validation regimes, and that interpretability and visualization are often prerequisites for adoption in clinical workflows~\cite{intro_3,trajvis2024jamia}. 
Visualization systems that translate trajectory models into clinician-facing interfaces illustrate the importance of closing the loop from model to decision, particularly for chronic disease management over time~\cite{trajvis2024jamia}. 
Consequently, a Nature-style positioning for new work should explicitly declare the trajectory object, the intervention object (sequence vs.\ regime/policy), the confounding and observation assumptions, and the decision/evaluation target, to avoid architectural novelty claims that do not resolve identifiability or data-support limitations~\cite{intro_1,2-1,2-17,feuerriegel2024natmed}. 

\section{Discrete-Time, Intervention-Aware Sequence Approaches}
\label{sec:discrete-intervention}
From observational discrete-time longitudinal data sliced by follow-up visits, clinical consultations and time windows, individualized response trajectories under distinct time-series treatment regimens are estimated~\cite{2-1,2-2,2-3}. The typical data structure is described as follows:
\[
\left\{ X_t, A_t, Y_{t+1} \right\}_{t=1}^T
\]
Here, $X_t$ denotes a vector of time-varying covariates, which include evolving clinical measurements such as laboratory test results and vital signs. The variable $A_t$ represents a possibly multivalued or multivariate time-varying treatment, which may change over time in response to the observed history. The outcome $Y_t$ denotes a time-varying response of interest, which may be modeled jointly with $X_t$ when covariates and outcomes are generated by a shared underlying process.

The inferential target is not restricted to the risk at a single terminal time point $H_t = \{ V, X_{\le t}, A_{< t}, Y_{\le t}\}$, but rather focuses on predicting the future potential outcome trajectory $Y^{a}_{t+1:t+\tau}$ over multiple steps or even the entire remaining horizon, under a hypothetical future treatment sequence $a_{t+1:t+\tau}$, conditional on the observed history at time $t$~\cite{2-2,2-3,2-4}. In this manner, the effects of alternative treatment sequences or policies can be compared, facilitating the development of individualized dynamic treatment regimes~\cite{2-1,2-4,2-5}.

The discrete-time literature can be summarized as a progression from (i) translating MSM/g-formula ideas into scalable deep sequence learners (e.g., RMSN, G-Net), which embed classical longitudinal causal adjustment into neural sequence modeling for time-varying treatments~\cite{robins2000msm,2-17,2-11,2-18}. 
It then moves to (ii) learning counterfactual states via balanced representations and attention (e.g., CRN and the Causal Transformer), aiming to reduce confounding by constructing history representations that are predictive for outcomes yet less predictive of treatment assignment~\cite{2-2,2-3}. 

A third phase emphasizes (iii) long-horizon robustness through empirical scrutiny and representation learning, including evidence that balancing is not uniformly beneficial (ICML 2024) and methods that use self-supervision/contrastive objectives to strengthen temporal representations (e.g., COSTAR and causal contrastive learning)~\cite{2-12,2-10,2-13}.
Most recently, the field has expanded toward (iv) moving beyond point forecasts to distributional counterfactuals and inference-oriented estimators, including counterfactual generative models for time-varying treatments and frameworks targeting uncertainty-aware or doubly robust estimation (e.g., Bayesian g-computation and TMLE-based estimators)~\cite{2-4,2-9,2-16}. Representative work and methodologies are summarized in Table~\ref{tab:sec3_discrete_methods}.

For disease trajectory studies, the practical takeaway is that method choice should be driven by the intended decision use (e.g., mean prediction versus risk-sensitive or distributional decision-making)~\cite{2-4,2-1}. 
It should also be guided by the severity of time-varying confounding and overlap/positivity limitations in the data, which can materially affect long-horizon rollouts~\cite{robins2000msm,2-17,2-12}. 
Finally, the horizon at which counterfactual rollouts must remain stable should be treated as a first-class design constraint, because errors and distribution shift can compound with time~\cite{2-12,2-11,2-3}. 
\begin{table*}[t]
\centering
\small
\setlength{\tabcolsep}{4pt}
\renewcommand{\arraystretch}{1.2}
\caption{Discrete-time intervention-aware sequence models.Representative families for estimating counterfactual outcome trajectories from visit-indexed observational sequences with time-varying treatments.}
\label{tab:sec3_discrete_methods}
\begin{xltabular}{\textwidth}{@{}p{0.21\textwidth}p{0.23\textwidth}X X@{}}
\toprule
\textbf{Method family} &
\textbf{Representative works} &
\textbf{Core contribution} &
\textbf{Key caveats / typical failure modes} \\
\midrule
\endfirsthead
\toprule
\textbf{Method family} &
\textbf{Representative works} &
\textbf{Core contribution} &
\textbf{Key caveats / typical failure modes} \\
\midrule
\endhead
\midrule
\multicolumn{4}{r}{\emph{Continued on next page}}\\
\endfoot
\bottomrule
\endlastfoot

\makecell{G-method foundations \\$\rightarrow$ deep sequence learners} &
\makecell[l]{MSMs/IPW~\cite{robins2000msm,2-17};\\ RMSN~\cite{2-11}; DSW~\cite{2-8}} &
Operationalizes longitudinal causal adjustment (IPW, $g$-methods) in scalable temporal models for time-varying treatments and outcomes &
Relies on sequential ignorability and overlap; long-horizon IPW can be high-variance; sensitive to model misspecification and treatment--covariate feedback~\cite{2-1,robins2000msm} \\

Balanced representations and attention for counterfactual forecasting &
Causal Transformer~\cite{2-3} &
Uses attention-based temporal representations and balancing objectives to improve long-range counterfactual prediction compared with RNN baselines &
Balancing does not guarantee identification under hidden confounding; gains can be benchmark-dependent; still constrained by overlap for long horizons~\cite{2-1} \\

Long-horizon representation robustness via contrastive objectives &
Causal contrastive learning~\cite{2-13} &
Uses contrastive predictive coding / mutual-information principles to improve temporal representation quality and stability for counterfactual regression &
Contrastive proxies may not align with downstream clinical utility; robustness depends on augmentation design and support in the logged data~\cite{2-13,2-1} \\

Distributional counterfactual trajectories (beyond point estimates) &
Counterfactual generative models~\cite{2-4} &
Enables distributional summaries (e.g., quantiles, tail risk) for safety-relevant decisions using conditional generative modeling with IPW-inspired correction &
Depends on correct treatment-assignment modeling and overlap; distributional realism is hard to validate under limited counterfactual support~\cite{2-4,2-1} \\
\hline

\end{xltabular}
\addtocounter{table}{-1}
Abbreviations: MSM, marginal structural model; IPW, inverse probability weighting.
\end{table*}

\subsection{Longitudinal Potential Outcomes and Time-Varying Confounding}
Most existing methods operate under a longitudinal extension of the Neyman-Rubin potential outcomes framework, in which causal effects are defined as comparisons between an individual’s potential outcomes under different interventions~\cite{2-1,2-2,2-3,2-13,2-4}. For each individual and each possible treatment history $\bar{a}_{1:T}$, a complete potential outcome trajectory $Y^{\bar{a}}_{1:T}$ is conceptually well defined, whereas in practice only a single factual trajectory corresponding to the realized treatment history $\bar{A}_{1:T}$ is observed, with all remaining potential trajectories being missing~\cite{2-1,2-2}. The central methodological challenge arises from time-varying confounding induced by treatment-confounder feedback, whereby current covariates $L_t \subset X_t$ are affected by past treatments $A_{<t}$ while simultaneously predicting both future treatment assignment $A_t$ and subsequent outcomes $Y_{>t}$~\cite{2-1,2-2,2-8}. As a consequence, directly applying black-box sequential models of the form $Y \sim f(H)$ often yields systematically biased estimates of intervention effects, since such models implicitly condition on post-treatment variables, including mediators and downstream consequences of prior treatments~\cite{2-1,2-2,2-8}.

Consequently, this class of methods must incorporate explicit causal adjustment mechanisms alongside predictive modeling. Representative approaches include longitudinal inverse probability weighting and marginal structural models, as well as the g-formula, which reweight or standardize over treatment assignment mechanisms to account for time-varying confounding~\cite{2-1,2-8,2-9}; representation learning methods combined with domain adversarial objectives or mutual information penalties, which aim to construct balanced latent representations in which treatment assignment is independent of confounders~\cite{2-2,2-3,2-10,2-13}; and approaches that explicitly model latent confounding factors or adjustment variables, thereby capturing unobserved sources of bias that jointly influence treatments and outcomes over time~\cite{2-8,2-9}.

\subsection{Discrete-time Modeling}

\label{discrete}
This class of methods discussed in this chapter almost universally adopts a discrete-time modeling framework~\cite{2-1,2-8}, in which time is partitioned into steps $t = 1, \ldots, T$ based on clinical visits, days or weeks, or fixed observation windows. At each time step, the tuple $(X_t, A_t, Y_t)$ is updated, and the data-generating process is modeled recursively conditional on the observed history $\{H_t\}$.

The discrete-time formulation offers several practical advantages. First, it allows direct reuse of mature sequential modeling architectures, such as recurrent neural networks (RNN)~\cite{rnn}, Transformers~\cite{transformer}, and state-space models~\cite{2-1,2-2,2-3}. Second, it enables the straightforward application of longitudinal g-methods in discrete time, including window-based marginal structural models with inverse probability weighting and g-computation, thereby facilitating a principled connection with traditional epidemiological approaches~\cite{2-1,2-8,2-9}.

\subsubsection{Learning counterfactual sequences: adjustment and long-horizon representations}
\label{lss}
Early deep sequence models in this space can be read as neural instantiations of these g-method ideas, designed to scale to high-dimensional EHR histories and long-horizon rollouts~\cite{intro_1,2-1}. Recurrent Marginal Structural Networks (RMSNs) explicitly draw inspiration from MSMs by learning recurrent models of treatment assignment and outcomes and using propensity-based weighting to reduce time-dependent confounding when forecasting responses to planned treatment sequences~\cite{2-11}. Deep Sequential Weighting (DSW) follows a similar philosophy—learning recurrent representations tied to treatment assignment and outcome evolution, then computing time-varying inverse probabilities used for sequential reweighting to estimate longitudinal individualized treatment effects~\cite{2-8}. These approaches represent one branch of the field: keep the epidemiologic correction mechanism (weights), but make both propensity estimation and outcome modeling flexible sequence learners~\cite{2-11,2-1,2-8}. 

A second branch replaces explicit weighting with balanced representation learning, aiming to make the learned state at time (t) predictive for outcomes but minimally informative about treatment assignment once history is encoded~\cite{2-2,2-3}. The Counterfactual Recurrent Network (CRN) is a canonical example: it uses an encoder–decoder RNN to map histories into a latent representation and to decode multi-step counterfactual outcomes under hypothetical future treatment plans, while applying a domain-adversarial objective to encourage treatment-invariant representations at each time step~\cite{2-2}. A generic form of the learning problem is a min–max objective,

\[
\min_{\theta,\phi} \max_{\psi}
\Bigg[
\sum_{t} \ell\big(Y_{t+1}, f_{\theta}(r_{\phi}(H_t), A_t)\big)
-
\lambda \sum_{t} \ell_{\mathrm{CE}}\big(A_t, d_{\psi}(r_{\phi}(H_t))\big)
\Bigg]
\]

where $r_{\phi}$ is the representation, $d_{\psi}$ is an adversary predicting treatment from the representation, and $\lambda$ controls the strength of balancing~\cite{2-2,2-3}.

The intuition is that if $r_{\phi}(H_t)$ “forgets” treatment-selection artifacts while retaining prognostic information, then decoding outcomes under alternative treatment sequences becomes closer to the desired counterfactual prediction~\cite{2-2,2-3}.

As datasets grew longer and confounding depended on long-range temporal patterns, attention-based architectures became attractive because they can model dependencies across many time steps without the inductive biases and optimization issues of recurrence~\cite{2-3}. The Causal Transformer (CT) adapts the transformer paradigm to temporal counterfactual estimation by processing time-varying covariates, past treatments, and past outcomes with dedicated attention modules and coupling them via cross-attention, while still using a “counterfactual domain confusion” loss to mitigate confounding bias~\cite{2-3}. In this sense, CT can be interpreted as carrying the CRN balancing idea into an architecture better suited for long-horizon dependence and complex temporal interactions~\cite{2-2,2-3}. 

However, a key recent lesson is that balancing is not a free lunch in the temporal setting. Huang et al.~\cite{2-12} conduct a large empirical study (ICML 2024) and show that adding balancing modules does not automatically improve multi-step counterfactual accuracy, and that plain empirical risk minimization can sometimes outperform balancing-augmented variants on temporal counterfactual benchmarks. This result matters for disease-trajectory applications because long-horizon rollouts amplify small modeling errors, and overly constraining representations toward invariance can trade away predictive signal needed for stable multi-step forecasts~\cite{2-12,2-1}. 

This empirical perspective has helped motivate a shift from “balancing as the main trick” toward “representation quality as the main bottleneck,” where self-supervision and contrastive objectives are used to learn more transferable temporal states before (or alongside) counterfactual supervision. COSTAR~\cite{2-10,2-13} integrates self-supervised learning into temporal counterfactual estimation, combining temporal/feature-wise attention with a component-wise contrastive loss to improve history representations and generalization under distribution shift~\cite{2-10}. A standard contrastive objective used in such settings is InfoNCE,
\[
\mathcal{L}_{\mathrm{NCE}}
=
-\log
\frac{
\exp\big(\mathrm{sim}(z, z^{+})/\tau\big)
}{
\exp\big(\mathrm{sim}(z, z^{+})/\tau\big)
+
\sum_{k=1}^{K}
\exp\big(\mathrm{sim}(z, z_k^{-})/\tau\big)
}
\]
which encourages representations $z$ to be close to "positive" views $z^{+}$ and far from negatives $\{z_k^{-}\}$~\cite{2-10}.

In parallel, Causal Contrastive Learning (NeurIPS 2024) shows that one can remain with an RNN backbone yet use contrastive predictive coding and InfoMax principles to better capture long-term dependencies for temporal counterfactual regression, explicitly positioning itself as a more efficient long-horizon alternative to transformer-heavy designs~\cite{2-13}.

\subsubsection{From prediction to decision: uncertainty, distributions and robustness}
\label{fptdu}
A third branch broadens the target from mean counterfactual trajectories to distributional counterfactual trajectories, arguing that clinical decisions may depend on risk, tail events, or heterogeneity that a point estimate cannot express. Wu et al. ~\cite{2-4} propose counterfactual generative modeling for time-varying treatments, training conditional generative models—such as conditional VAEs and guided diffusion—to sample full counterfactual outcome trajectories under alternative treatment sequences~\cite{2-4}. A key idea is to handle the mismatch between the observed factual distribution and the desired counterfactual distribution through an IPTW-inspired loss, rather than requiring explicit tractable likelihoods for the entire longitudinal process~\cite{2-4,robins2000msm}. This line of work connects the “trajectory as rollout under intervention” view with modern generative modeling, and it naturally supports uncertainty-aware summaries (e.g., quantiles, worst-case risk) that align with safety-sensitive clinical decision-making~\cite{2-4,2-1}. 

An underdeveloped but increasingly important direction in this branch is \emph{medical-knowledge augmentation}: injecting ontologies, medical knowledge graphs, and clinically curated relation priors into temporal counterfactual models so that latent states are not only predictive, but also semantically grounded for interpretation and decision support~\cite{intro_1,choi2017gram,yang2023path,wuuhar}. Instead of learning solely from raw co-occurrence statistics in EHR sequences, a knowledge graph $\mathcal{G}=(\mathcal{V},\mathcal{E})$ can provide structured constraints over diagnoses, medications, procedures, and their typed relations. A common encoder in this setting uses graph message passing,
\[
H^{(\ell+1)}=\sigma\!\left(\tilde D^{-1/2}\tilde A\tilde D^{-1/2}H^{(\ell)}W^{(\ell)}\right),
\]
where $\tilde A=A+I$ and $\tilde D$ is its degree matrix; node embeddings from the final layer are then aligned to patient histories as concept-level priors for downstream trajectory estimation~\cite{choi2017gram,yang2023path}.

For longitudinal counterfactual prediction at time $t$, this is often implemented as a fusion map
\[
z_t = \phi\!\left([h_t^{\mathrm{EHR}}\,\|\,h_t^{\mathrm{KG}}]\right),
\qquad
\hat y_{t+1}^{(a)} = f_{\theta}(z_t,a_t),
\]
where $h_t^{\mathrm{EHR}}$ is the temporal representation from raw observations and $h_t^{\mathrm{KG}}$ is the knowledge-guided representation derived from ontology/KG neighborhoods. To preserve medical consistency and improve explainability, many implementations add a graph smoothness regularizer
\[
\mathcal L
=
\mathcal L_{\mathrm{cf}}
+\lambda\!\sum_{(u,v)\in\mathcal E} w_{uv}\,\lVert z_u-z_v\rVert_2^2,
\]
which penalizes clinically implausible representation jumps between strongly related medical concepts~\cite{choi2017gram,2-3,2-10}. From a decision-support perspective, this matters because treatment recommendations can be traced to medically meaningful concept paths rather than opaque latent dimensions, improving auditability when models are deployed in high-stakes workflows~\cite{yang2023path,feuerriegel2024natmed,gottesman2019rlhc}.

At the systems level, knowledge augmentation can also improve feature understanding of sparse raw data by anchoring rare codes to higher-level clinical abstractions (e.g., disease hierarchies, drug classes), which is especially useful under small-sample regimes and site shift~\cite{rotmensch2017hkg,rasmy2021,li2020behrt}. However, Nature-grade evidence standards require that these gains be evaluated beyond accuracy deltas: reports should test whether knowledge-enhanced models improve calibration, counterfactual plausibility, and policy robustness under overlap stress, rather than only showing better AUROC on observed-care prediction tasks~\cite{feuerriegel2024natmed,hernan2016targettrial,2-17}. 

Latent-variable models provide another route to richer counterfactuals, especially when unobserved heterogeneity or hidden factors drive outcome dynamics~\cite{2-15,2-8}. CDVAE ~\cite{2-15} introduces a dynamic variational autoencoder perspective for longitudinal counterfactual regression, using latent adjustment variables intended to capture unobserved risk factors affecting outcome sequences while supporting counterfactual effect estimation over time. At a high level, VAE-based trajectory modeling optimizes an evidence lower bound (ELBO) such as
\[
\log p_\theta(y \mid c)
\ge
\mathbb{E}_{q_\phi(z \mid y,c)}
\left[
\log p_\theta(y \mid z,c)
\right]
-
\mathrm{KL}
\big(
q_\phi(z \mid y,c)
\,\|\, 
p(z)
\big)
\]

where $c$ bundles observed history and treatments and $z$ is the latent variable~\cite{2-15}.
DSW also explicitly frames part of the problem as learning latent representations tied to treatment assignment, then uses sequential weighting to adjust for time-varying confounding~\cite{2-8}.

Finally, some discrete-time approaches emphasize inference and decision validity rather than purely predictive accuracy~\cite{2-9,2-16}. Xu et al.~\cite{2-9} adapt Bayesian g-computation using multivariate generalized linear mixed-effects models to jointly model outcomes, time-varying confounders, and treatment assignment with subject-specific random effects, enabling posterior uncertainty over counterfactual trajectories under dynamic regimes. DeepNetTMLE~\cite{2-16} integrates deep longitudinal representation learning with targeted maximum likelihood estimation (TMLE) under general interference, aiming to retain doubly robust, inference-oriented properties while modeling time-varying confounding in networked settings. These approaches underscore an important conclusion: for intervention-aware trajectory models to be decision-support tools, we often need calibrated uncertainty, robustness to overlap violations, and estimators whose errors are less sensitive to nuisance model misspecification~\cite{2-9,2-16,robins2000msm}. 

\section{Continuous-Time, Irregular, and Policy-Level Modeling}
\label{sec:continuous-policy}

Continuous-time, intervention-aware trajectory modeling emerged because the time axis in healthcare is not a bookkeeping detail but part of the data-generating mechanism~\cite{intro_1,pmlr-v202-vanderschueren23a}.
Measurements and treatments occur asynchronously, and their timestamps reflect both clinical workflows and evolving patient severity~\cite{intro_1,che2018missing}. 
As a result, discretizing trajectories into fixed bins is not merely a convenience; it can destroy clinically meaningful timing information, introduce preprocessing artifacts, and implicitly merge the disease process with the observation process~\cite{intro_1,kidger2020ncde,pmlr-v202-vanderschueren23a}. 

This becomes especially problematic when the scientific goal shifts from forecasting ("what happens next?") to counterfactual reasoning ("what would have happened under different treatment decisions?")~\cite{2-1,2-17}. 
Treatment assignment is history-dependent, so purely predictive training on logged trajectories can encode confounding-induced correlations rather than causal effects~\cite{2-1,2-17}. 
Continuous-time methods are therefore best understood as a response to two joint pressures: irregular data streams that resist grid-based modeling and causal questions that require well-defined counterfactual outcomes at arbitrary horizons~\cite{pmlr-v162-seedat22b,pmlr-v236-ying24a}. 

At the same time, the field has increasingly moved from “sequence-level” interventions (setting a fixed future treatment path) toward “policy-level” interventions (changing the rule that generates treatments)~\cite{schulam2017reliable,pmlr-v202-hizli23a}. 
This matters because clinical decisions are typically adaptive, driven by new laboratory results, symptoms, and clinician beliefs, so interventions on the policy change the \emph{distribution} over future treatment times and doses rather than just one realized sequence~\cite{pmlr-v202-hizli23a,gottesman2019rlhc}.
This methodology sits precisely at this intersection: continuous time, irregular sampling, and policy semantics~\cite{intro_1,2-1}. Representative work and methodologies are summarized in Table~\ref{tab:sec4_continuous_methods}.

\begin{table*}[t]
\centering
\small
\setlength{\tabcolsep}{4pt}
\renewcommand{\arraystretch}{1.2}
\caption{Continuous-time, irregular, and policy-level modeling. Representative method families for irregularly sampled trajectories, continuous-time counterfactuals, observation-process correction, and policy-level interventions.}
\label{tab:sec4_continuous_methods}
\begin{xltabular}{\textwidth}{@{}p{0.22\textwidth}p{0.23\textwidth}X X@{}}
\toprule
\textbf{Method family} &
\textbf{Representative works} &
\textbf{Core contribution} &
\textbf{Key caveats / typical failure modes} \\
\midrule
\endfirsthead
\toprule
\textbf{Method family} &
\textbf{Representative works} &
\textbf{Core contribution} &
\textbf{Key caveats / typical failure modes} \\
\midrule
\endhead
\midrule
\multicolumn{4}{r}{\emph{Continued on next page}}\\
\endfoot
\bottomrule
\endlastfoot

Irregular-time backbones (from time-aware RNNs to neural DEs) &
T-LSTM~\cite{baytas2017tlstm}; GRU-D~\cite{che2018missing}; Neural ODE~\cite{chen2018neuralode}; Neural CDE~\cite{kidger2020ncde} &
Provides principled representations for irregular sampling, separating modeling of temporal dynamics from arbitrary binning and enabling prediction at arbitrary times &
Backbone expressivity alone does not confer causal validity; observation processes can be informative and confounded with severity~\cite{intro_1} \\

Continuous-time counterfactual trajectories via neural differential equations &
TE-CDE~\cite{pmlr-v162-seedat22b}; CF-ODE~\cite{pmlr-v151-de-brouwer22a} &
Extends counterfactual trajectory estimation to continuous time and supports arbitrary horizons and uncertainty-aware decision support components &
Requires continuous-time analogues of sequential ignorability and overlap; counterfactual extrapolation is fragile without uncertainty quantification and strong evaluation~\cite{2-1,pmlr-v151-de-brouwer22a} \\

Observation-process correction for informative sampling &
TESAR-CDE~\cite{pmlr-v202-vanderschueren23a} &
Treats visit timing as part of the data-generating process and corrects forecasting bias via inverse-intensity weighting for observation events &
Needs credible modeling of observation intensities; informative sampling can remain partially unmeasured; weights can be unstable for long horizons~\cite{pmlr-v202-vanderschueren23a} \\

Policy-level interventions (intervening on treatment rules) &
Policy intervention modeling~\cite{pmlr-v202-hizli23a}; counterfactual decision support~\cite{schulam2017reliable} &
Formalizes counterfactual reasoning about policies in continuous time, aligning with adaptive clinical decision-making and enabling policy simulation &
OPE is highly sensitive to overlap and model error; policy changes induce distribution shift in both actions and observation patterns~\cite{gottesman2019rlhc,pmlr-v202-hizli23a} \\

OPE and RL cautionary baselines for clinical policy claims &
OPE critique in sepsis~\cite{pmlr-v68-raghu17a}; AI Clinician~\cite{komorowski2018aiclinician}; RL guidelines~\cite{gottesman2019rlhc} &
Clarifies failure modes of policy-learning/OPE in healthcare (confounding, non-stationarity, poor overlap), motivating uncertainty-aware and conservative evaluation &
Clinical deployment requires careful causal assumptions, auditing, and external validation; optimistic OPE can be misleading under weak support~\cite{gottesman2019rlhc,pmlr-v68-raghu17a} \\

Multi-agent dynamics and interference in continuous time &
Causal Graph ODE (CAG-ODE)~\cite{huang2024cagode} &
Introduces a framework for counterfactual modeling with interference using graph-structured neural ODEs and causal objectives &
Interference assumptions and network measurement are often violated in practice; scalable evaluation remains challenging~\cite{huang2024cagode} \\

\hline
\end{xltabular}
\addtocounter{table}{-1}
Abbreviations: ODE/CDE, (controlled) ordinary differential equation; OPE, off-policy evaluation; RL, reinforcement learning.
\end{table*}

\subsection{Continuous-time causal targets: paths, policies, and the observation process}

A useful abstraction is to treat patient data as coupled processes over t in [0, T]: covariates X(t), treatments A(t), outcomes Y(t), and an observation/visit process O(t) that determines when measurements are recorded~\cite{intro_1,pmlr-v202-vanderschueren23a}. 
In EHR-style datasets, we only observe these processes at irregular event times, making “the history” a sigma-field that includes both values and timestamps ~\cite{intro_1,pmlr-v202-vanderschueren23a}.

For intervention-aware trajectory modeling, a canonical estimand is the expected
potential outcome at time $t$ under a treatment path $a(\cdot)$, e.g.,
$\mathbb{E}[Y^{a}(t) \mid X(0)=x_0]$, where the counterfactual trajectory is defined
for any $t$ rather than only for discrete steps~\cite{pmlr-v162-seedat22b,hess2024scipnet}.

Policy-level targets instead consider a (possibly stochastic) policy $\pi$ that maps
history to treatment decisions, producing counterfactual outcomes $Y^{\pi}(t)$ and
objectives such as $\mathbb{E}[Y^{\pi}(t)]$ or long-horizon value criteria~\cite{pmlr-v202-hizli23a,gottesman2019rlhc,pmlr-v236-ying24a}.

Identification in continuous time relies on analogues of consistency, sequential ignorability, and positivity/overlap, but these assumptions must be interpreted carefully because treatment-confounder feedback can occur at arbitrarily fine resolution when monitoring is dense~\cite{pmlr-v236-ying24a,2-17}. 
This is one reason counting-process representations are natural: treatment changes and observation events can be modeled via intensities, and adjustment can be framed through continuous-time likelihood ratios or reweighting arguments rather than fixed-step propensity scores~\cite{hess2024scipnet,pmlr-v236-ying24a}. 
Empirically, most continuous-time causal ML models instantiate these ideas through some combination of learned balancing representations and explicit inverse-propensity or inverse-intensity weighting adapted to irregular timestamps~\cite{pmlr-v162-seedat22b,pmlr-v202-vanderschueren23a,hess2024scipnet}.

\subsection{Representational backbones for irregular time: from "time-aware RNNs" to neural ODE/CDE/SDE}

\label{rep}

 Historically, irregularity was first handled within discrete-time sequence models by injecting time-gap features and decay mechanisms, which improved prediction but still treated trajectories as sequences of updates rather than as objects defined continuously in time~\cite{baytas2017tlstm,che2018missing}. 
  Time-aware LSTMs (e.g., T-LSTM) decayed memory as a function of elapsed time, while GRU-D explicitly modeled informative missingness patterns via decay-to-mean and masking/time-interval inputs~\cite{baytas2017tlstm,che2018missing}. These approaches can be effective, but they do not inherently provide a principled notion of patient state at arbitrary timestamps without additional interpolation rules. 

  Neural differential equations offered a conceptual shift: learn a continuous-time dynamical system that can be integrated forward to any time, rather than a discrete transition map between bins~\cite{chen2018neuralode}. 
  
  In a neural ODE, a latent state $z(t)$ evolves via
$\frac{d z(t)}{d t} = f_\theta(z(t), t)$,
and predictions at arbitrary $t$ are obtained by numerical integration and a readout
$\hat{y}(t) = g_\phi(z(t))$~\cite{chen2018neuralode}.

 Neural CDEs further align with the structure of irregular multivariate time series
by treating the observed (interpolated) data stream $u(t)$ as a control that drives
$z(t)$ through an integral equation of the form
$z(t) = z(0) + \int_0^t f_\theta(z(s))\, d u(s)$~\cite{kidger2020ncde}.

  This yields a principled continuous-time analogue of how RNNs process streams, while making explicit the modeling choices around interpolation and path representation~\cite{kidger2020ncde}.
  Neural SDE variants then introduce stochasticity to represent process noise and support uncertainty quantification through posterior predictive distributions rather than point trajectories~\cite{hess2023bncde}. 

 \subsection{Making continuous-time models causal: balancing, weighting, uncertainty, and hidden structure}
\label{mctm}
  The methodological leap from “continuous-time forecasting” to “continuous-time counterfactuals” requires two elements: (i) treatments must be represented as part of the driving signal or event process, and (ii) time-varying confounding must be addressed to justify causal interpretation~\cite{2-1,pmlr-v162-seedat22b}.
  
  TE-CDE illustrates a balancing-driven approach: it uses a neural CDE latent state to estimate counterfactual outcome trajectories at arbitrary horizons under irregular sampling, while using adversarial balancing to reduce dependence between treatment assignment and latent history~\cite{pmlr-v162-seedat22b}. 
  This can be viewed as a continuous-time analogue of discrete-time balanced-representation causal sequence models, but with dynamics defined over real time rather than over bins~\cite{pmlr-v162-seedat22b,2-2}. 

  Another theme is that counterfactual prediction is fundamentally an extrapolation problem, so uncertainty is not optional if the goal is reliable decision support~\cite{pmlr-v151-de-brouwer22a,gottesman2019rlhc}.
  CF-ODE is representative here, combining an ODE-based latent dynamics model with uncertainty-aware inference to highlight when counterfactual trajectories are unsupported due to weak overlap or distribution shift between factual and counterfactual regimes~\cite{pmlr-v151-de-brouwer22a}. 
  Bayesian neural CDE/SDE approaches provide a complementary route by producing posterior predictive distributions for counterfactual trajectories, enabling downstream decision systems to incorporate epistemic uncertainty~\cite{hess2023bncde,pmlr-v151-de-brouwer22a}. 

  In contrast, weighting-based continuous-time methods aim to stay close to classical g-method logic by explicitly reweighting observed trajectory segments to emulate a target intervention distribution~\cite{hess2024scipnet,pmlr-v236-ying24a}. 
  SCIP-Net is notable because it derives stabilized inverse-propensity weighting in continuous time and integrates these stabilized weights into neural estimation of conditional average potential outcomes under irregular treatment and measurement times~\cite{hess2024scipnet}. 
  The underlying message is classical but amplified by irregular time: without stabilization, long-horizon products of likelihood ratios can yield extreme weights and high-variance estimation, even if the representation backbone is expressive~\cite{hess2024scipnet,2-17}. 

  Importantly, irregularity is not only about treatment times; it is also about \emph{when we observe outcomes and covariates}, and those timestamps are often informative~\cite{intro_1,pmlr-v202-vanderschueren23a}. 
  If sicker patients are monitored more frequently, then naive training on observed samples can induce bias in both trajectory prediction and treatment-effect estimation~\cite{pmlr-v202-vanderschueren23a}. 
  TESAR-CDE makes this explicit by modeling informative sampling as covariate shift and using inverse-intensity weighting for the observation process, instantiated with a neural CDE forecaster~\cite{pmlr-v202-vanderschueren23a}. 
  This perspective reframes “irregular sampling” as a causal problem: without modeling or correcting the observation process, continuous-time models can remain biased regardless of their dynamical expressivity~\cite{pmlr-v202-vanderschueren23a,intro_1}. 

  Hidden confounding remains a central unresolved issue, because latent severity can drive both treatment escalation and outcomes, compounding bias over time~\cite{2-1,2-17}. 
  In discrete time, the Time Series Deconfounder exemplifies proxy/latent-factor strategies that infer latent confounding structure from multi-treatment patterns~\cite{pmlr-v119-bica20a}. 
  In continuous time, LipCDE and extensions combine continuous-time dynamics with regularization constraints and latent-factor modeling to mitigate hidden confounding under irregular observations~\cite{caoEWSMNL23,cao2023ct_hidden}. 
  These approaches are promising, but the field still lacks standardized continuous-time benchmarks and sensitivity analysis protocols that would make robustness claims comparable across methods ~\cite{2-1,caoEWSMNL23}.

  \subsection{Policy-level and multi-agent modeling: from “what-if trajectories” to decision-making}
\label{plmam}
  Sequence-level counterfactuals answer “what if the future treatment path were a(t)?,” but real clinical questions often ask “what if the policy were changed?” because future actions are adaptive and sometimes stochastic~\cite{schulam2017reliable,pmlr-v202-hizli23a}. 
  Policy-level modeling therefore aims to represent the treatment assignment mechanism itself, often as a temporal point process, and to support interventions that modify that mechanism rather than setting one deterministic trajectory~\cite{pmlr-v202-hizli23a}. 

  Hızlı et al. are representative of this direction by coupling Gaussian process outcome trajectories with a point-process model for treatment events, enabling counterfactual simulation under policy interventions~\cite{pmlr-v202-hizli23a}. 
  In simplified terms, if treatment events follow a counting process with intensity
$\lambda_A^{\pi}(t \mid H(t))$ under policy $\pi$, then a policy intervention can be conceptualized as replacing $\lambda_A^{\pi}$ with $\lambda_A^{\pi'}$ while leaving the rest of the system fixed~\cite{pmlr-v202-hizli23a}.
  This aligns with earlier counterfactual decision-support perspectives that emphasize that predictive accuracy under the behavior policy does not guarantee reliable decision support under alternative actions~\cite{schulam2017reliable}. 

  The policy-level viewpoint also clarifies why offline evaluation is fragile: it requires overlap and credible counterfactual modeling, and can fail catastrophically when models extrapolate beyond logged support~\cite{gottesman2019rlhc,pmlr-v151-de-brouwer22a}. This is precisely the motivation behind reinforcement-learning guidelines for healthcare and the critical scrutiny of ICU policy-learning case studies in sepsis~\cite{gottesman2019rlhc,pmlr-v68-raghu17a,komorowski2018aiclinician}. 
  From a trajectory-modeling perspective, continuous-time counterfactual models can function as model-based components for policy evaluation, but only if uncertainty, overlap, and confounding are treated as first-order concerns rather than afterthoughts~\cite{pmlr-v151-de-brouwer22a,gottesman2019rlhc}. 

  Finally, many clinically relevant settings involve interference, where one unit’s treatment affects another’s outcomes (e.g., infectious disease dynamics or spatial policy)~\cite{2-16,huang2024cagode}. 
  CAG-ODE illustrates how graph-structured neural ODEs can be combined with treatment representations and balancing objectives to model continuous-time counterfactual dynamics in multi-agent systems~\cite{huang2024cagode}. 
  DeepNetTMLE provides a complementary line by integrating deep temporal models with targeted maximum likelihood estimation to support doubly robust inference under general interference~\cite{2-16}.
  These directions are early, but they begin to align continuous-time counterfactual modeling with real healthcare settings where “no interference” is often untenable~\cite{2-16,huang2024cagode}.

  \subsection{Bottlenecks and the evaluation agenda}

  The central bottleneck remains identifiability. Expressive dynamics do not by themselves confer causal validity, which depends on correct assumptions about confounding, overlap, and increasingly, observation processes~\cite{pmlr-v236-ying24a,2-17}.
Balancing objectives can be helpful inductive biases, but they do not guarantee identification under misspecification or hidden confounding.
Weighting approaches provide principled links to classical causal inference, but they inherit the difficulties of positivity and variance, which are further compounded by irregular event times and long-horizon weighting.

  Evaluation is the second major gap: many studies still rely on pointwise error metrics at selected horizons and on synthetic or semi-synthetic environments, while trajectory-level calibration, distributional accuracy, and decision-centric off-policy validation remain comparatively underdeveloped~\cite{intro_1,2-1,gottesman2019rlhc}. 
  This gap is especially consequential for policy-level claims, where the relevant question is not “how well do we predict logged outcomes?” but “how reliably do we evaluate or improve a policy under uncertainty and limited overlap?” ~\cite{gottesman2019rlhc,pmlr-v202-hizli23a}

  A longer-term methodological objective is therefore integration: unified dynamic causal models that jointly represent latent disease progression, treatment and observation intensities, uncertainty, and (when needed) interference, while remaining scalable and clinically interpretable~\cite{intro_1,2-1}.
  
  $ $\\

\section{Trajectory Modeling Without Explicit Causal Semantics}
\label{subsec:trajectory-without-causal-semantics}

This section is intentionally placed after the intervention-aware causal chapters. The preceding sections establish what is required to support counterfactual and policy-level claims---explicit treatment assignment assumptions, time-varying confounding control, positivity/overlap diagnostics, and observation-process modeling~\cite{2-1,2-17}. Only after that causal benchmark is clear can we interpret non-causal trajectory models as a principled boundary case: often highly valuable for forecasting and representation learning, but not sufficient on their own for intervention claims~\cite{intro_1,carrascoRibelles2023jamia_review}. This ordering therefore prevents semantic overreach and clarifies which questions these methods can answer reliably in clinical deployment.

\phantomsection
\label{tab:sec5_noncausal_anchor}

\begin{table*}[t]
\centering
\small
\setlength{\tabcolsep}{4pt}
\renewcommand{\arraystretch}{1.2}
\caption{Trajectory modeling without explicit causal semantics. Representative prognostic and generative trajectory families that model observed longitudinal evolution without encoding treatments as interventions via a counterfactual estimand.}
\label{tab:sec5_noncausal_methods}
\begin{xltabular}{\textwidth}{@{}p{0.23\textwidth}p{0.23\textwidth}X X@{}}
\toprule
\textbf{Method family} &
\textbf{Representative works} &
\textbf{Core contribution} &
\textbf{Key caveats / typical failure modes} \\
\midrule
\endfirsthead
\toprule
\textbf{Method family} &
\textbf{Representative works} &
\textbf{Core contribution} &
\textbf{Key caveats / typical failure modes} \\
\midrule
\endhead
\midrule
\multicolumn{4}{r}{\emph{Continued on next page}}\\
\endfoot
\bottomrule
\endlastfoot

Multi-state survival and counting-process formulations &
Aalen--Johansen~\cite{aalen1978aj}; counting-process framework~\cite{andersen1993counting}; multi-state tutorial~\cite{putter2007tutorial_multistate} &
Provides interpretable state-transition trajectories and principled handling of censoring and competing risks for dynamic prognosis &
State definition and state-space size can be limiting; conditioning on treatment-as-covariate remains associational unless causal structure is specified~\cite{2-1} \\

Joint longitudinal--event modeling for dynamic prediction &
Joint modeling foundations~\cite{wulfsohn1997joint_biometrics,rizopoulos2012joint_models_book} &
Supports dynamic prediction with explicit uncertainty and interpretable biomarker--event associations, widely used in clinical studies &
Scaling to high-dimensional EHR covariates is nontrivial; misspecification of longitudinal submodels or hazard links can degrade calibration~\cite{rizopoulos2012joint_models_book} \\

Deep sequence prediction on EHR "patient journeys" &
Doctor AI~\cite{choi2016doctorai}; RETAIN~\cite{choi2016retain}; large-scale deep EHR prediction~\cite{rajkomar2018}; Deep Patient~\cite{miotto2016deeppatient} &
Learns high-capacity longitudinal representations for forecasting and clinical risk stratification, with interpretable attention variants &
Forecasts reflect logged care and confounding; limited transparency in how representations map to mechanistic disease states~\cite{carrascoRibelles2023jamia_review} \\

Pretrained EHR transformers for longitudinal representation learning &
BEHRT~\cite{li2020behrt}; Med-BERT~\cite{rasmy2021} &
Improves transferability and performance across downstream tasks via pretraining on large EHR corpora &
Pretraining objectives may not align with clinical decision questions; robustness across sites and temporal shifts remains an open concern~\cite{carrascoRibelles2023jamia_review} \\

Missingness-aware and irregular-time prognostic sequence models &
GRU-D~\cite{che2018missing} &
Incorporates time gaps and missingness indicators to improve predictive performance under real EHR sampling patterns &
Still associational; informative sampling driven by care processes can bias downstream conclusions if used for decision support~\cite{intro_1} \\

Event-time modeling with temporal point processes &
Hawkes processes~\cite{hawkes1971}; RMTPP~\cite{du2016rmttp}; Neural Hawkes~\cite{mei2017neural_hawkes} &
Models the joint distribution of event types and times; supports simulation and interpretable excitation/interaction patterns &
Clinical event coding is noisy and affected by care pathways; linking intensities to latent disease state is nontrivial~\cite{intro_1} \\

Latent continuous-time trajectory models (GP-based and embeddings) &
Multi-resolution trajectories~\cite{schulam2015multiresolution}; Disease Trajectory Maps~\cite{schulam2016dtm} &
Captures heterogeneity with explicit uncertainty, supports subtyping/visualization, and handles irregular sampling naturally &
Latent factors may be weakly identifiable; clinical interpretability often requires additional supervision or post hoc analysis~\cite{intro_1} \\
\hline
\end{xltabular}
\addtocounter{table}{-1}

\end{table*}

Trajectory modeling \emph{without explicit causal semantics} refers to methods that learn and forecast longitudinal disease evolution, including events, states, biomarkers, and visit patterns, from observed data, without encoding treatments as interventions through a formal counterfactual estimand (e.g., potential outcomes under hypothetical treatment regimes)~\cite{intro_1,2-1}.

This chapter is central in practice because much of deployed trajectory AI remains anchored in predictive forecasting and dynamic risk updating under \emph{observed care}, rather than in identification of policy effects under hypothetical interventions~\cite{intro_1,carrascoRibelles2023jamia_review}.

\subsection{Scope and estimand boundary under observed care}
The common estimand in non-causal trajectory modeling is predictive, not interventional. For patient $i$ with history $\mathcal{H}_i(t)$ observed up to time $t$, models target a calibrated distribution over future observations on horizon $(t,t+\tau]$,
\begin{equation}
  p_{\theta}\!\left(\mathbf{O}_i\big((t,t+\tau]\big)\,\middle|\,\mathcal{H}_i(t)\right),
  \label{eq:predictive-trajectory}
\end{equation}
and derived quantities such as multi-step event risk, state occupation probabilities, or dynamic survival forecasts~\cite{putter2007tutorial_multistate,rizopoulos2012joint_models_book}.
For discretized horizons, this is often factorized autoregressively as
\begin{equation}
  p_{\theta}(\mathbf{O}_{i,t+1:t+H}\mid \mathcal{H}_i(t))
  =\prod_{h=1}^{H} p_{\theta}(O_{i,t+h}\mid \mathcal{H}_i(t+h-1)).
  \label{eq:autoregressive-factorization}
\end{equation}
The boundary is conceptual but crucial: conditioning on observed treatments in $\mathcal{H}_i(t)$ does not convert the predictive estimand in \hyperref[eq:predictive-trajectory]{Eq.~(\ref*{eq:predictive-trajectory})} into a counterfactual statement about intervention effects, especially under treatment--confounder feedback~\cite{2-1,2-17}.

\subsection{Classical probabilistic trajectory models: states, hazards, and dynamic prediction}
The most interpretable non-causal trajectory families begin with explicit stochastic-process structure~\cite{putter2007tutorial_multistate,andersen1993counting}. In multi-state formulations, disease evolution is represented by $S_i(t)\in\{1,\dots,K\}$ with intensity matrix $Q(t)$; transition probabilities $P(s,t)$ satisfy
\begin{equation}
  \frac{d}{dt}P(s,t)=P(s,t)\,Q(t),
  \label{eq:kolmogorov-forward}
\end{equation}
which yields clinically interpretable state occupancy and transition trajectories across follow-up~\cite{andersen1993counting,putter2007tutorial_multistate}. Under censoring, the Aalen--Johansen estimator remains a foundational nonparametric reference and a strong baseline for modern augmentations~\cite{aalen1978aj,putter2007tutorial_multistate}.

Joint longitudinal--event models provide a complementary framework for dynamic prediction by coupling a latent biomarker process to event hazard~\cite{wulfsohn1997joint_biometrics,rizopoulos2012joint_models_book}:
\begin{align}
  y_i(t_{ij}) &= m_i(t_{ij}) + \varepsilon_{ij}, \qquad \varepsilon_{ij}\sim\mathcal{N}(0,\sigma^2), \label{eq:jm-long}\\
  m_i(t) &= \mathbf{x}_i(t)^{\top}\beta + \mathbf{z}_i(t)^{\top}b_i, \qquad b_i\sim\mathcal{N}(0,\Sigma_b), \label{eq:jm-mixed}\\
  \lambda_i(t\mid b_i) &= \lambda_0(t)\exp\!\Big(\gamma^{\top}\mathbf{w}_i + \alpha\, m_i(t)\Big). \label{eq:jm-hazard}
\end{align}
Because posterior updates of $m_i(t)$ propagate directly into risk forecasts, these models remain among the clearest probabilistic accounts of trajectory updating in clinical studies, even though scaling to high-dimensional EHR covariates remains challenging~\cite{rizopoulos2012joint_models_book,eskofier2023deterioration_review}.

\subsection{Deep sequence forecasting and representation learning}
As EHR scale increased, trajectory modeling shifted toward high-capacity sequence encoders that optimize predictive likelihood under logged care pathways~\cite{intro_1,carrascoRibelles2023jamia_review}. RNN and attention architectures established the empirical template for next-visit and short-horizon trajectory forecasting~\cite{choi2016doctorai,choi2016retain}, while large EHR deep-learning pipelines and pretrained transformers demonstrated transfer gains across tasks~\cite{rajkomar2018,li2020behrt,rasmy2021}.

A generic training objective is
\begin{equation}
\mathcal{L}_{\mathrm{pred}}(\theta)
=-
\sum_{i,t}
\log p_{\theta}\!\left(Y_{i,t+1}\mid\mathcal{H}_i(t)\right),
\label{eq:pred-loss}
\end{equation}
possibly extended to multi-task heads for diagnoses, medications, readmissions, or mortality. This design is operationally effective, but its semantic target remains associational trajectory forecasting unless explicit causal estimands are added downstream~\cite{2-1,carrascoRibelles2023jamia_review}.

\subsection{Irregular-time, event-centric, and generative trajectories}
Irregular and informative observation is a recurrent bottleneck in real EHR data, because measurement times and missingness patterns reflect clinical decision processes rather than passive sampling~\cite{intro_1,che2018missing,eskofier2023deterioration_review}. Missingness-aware architectures such as GRU-D improve predictive performance by encoding time gaps and masks, but they still estimate observed-care prognosis, not intervention response~\cite{che2018missing,2-1}.

For event-centric trajectories, temporal point processes model both event type and timing via conditional intensity functions. For marked events $\{(t_n,m_n)\}_{n=1}^{N}$ on $[0,T]$, a standard likelihood is
\begin{equation}
  \log p(\{(t_n,m_n)\}_{n=1}^{N})
  =\sum_{n=1}^{N}\log \lambda_{m_n}(t_n\mid \mathcal{H}(t_n))
  -\int_{0}^{T}\sum_{k=1}^{K}\lambda_k(t\mid \mathcal{H}(t))\,dt,
  \label{eq:pp-loglik}
\end{equation}
which underlies Hawkes and neural point-process variants used for patient-journey simulation and forecasting~\cite{hawkes1971,mei2017neural_hawkes,du2016rmttp}.

Latent generative trajectory models extend this by learning hidden temporal states and sampling future trajectories with uncertainty. A core objective is the ELBO,
\begin{equation}
  \log p_{\theta}(\mathbf{y}_i)
  \ge
  \mathbb{E}_{q_{\phi}(z_i\mid \mathbf{y}_i)}\!\left[\log p_{\theta}(\mathbf{y}_i\mid z_i)\right]
  -\mathrm{KL}\!\left(q_{\phi}(z_i\mid \mathbf{y}_i)\,\|\,p(z_i)\right),
  \label{eq:elbo}
\end{equation}
with continuous-time latent dynamics often parameterized by neural ODE/CDE families to accommodate irregular timestamps~\cite{kingma2014vae,chen2018neuralode,kidger2020ncde,rubanova2019latentode}.

\subsection{Translational value, failure modes, and forward interface to causal modeling}
The principal limitation of this chapter is not predictive utility but claim scope. These models can be excellent for dynamic prognosis, subtype discovery, and uncertainty-aware simulation under observed care, yet still fail as decision engines when queried with intervention-style questions~\cite{2-1,intro_1}. Endogenous treatment assignment, informative observation, and policy drift can all induce stable predictive performance but unstable counterfactual behavior~\cite{2-1,2-17,carrascoRibelles2023jamia_review}.

Three recurring failure modes deserve explicit reporting: (i) semantic overreach, where associational forecasts are interpreted as treatment effects; (ii) transport fragility, where code practices and care pathways shift across institutions; and (iii) uncertainty under-specification, where point forecasts mask extrapolation outside empirical support~\cite{intro_1,wiens2019donoHarm,feuerriegel2024natmed,pmlr-v151-de-brouwer22a}. Nevertheless, non-causal trajectory models remain indispensable: they provide calibrated baselines, interpretable state summaries, and robust representation modules that form the practical substrate for subsequent intervention-aware pipelines~\cite{intro_1,eskofier2023deterioration_review,putter2007tutorial_multistate}. Representative method families are summarized in Table~\ref{tab:sec5_noncausal_methods}.

\section{Discussion}
\label{sec:discussion}

\begin{figure}[t]
 \centering
 \includegraphics[width=\linewidth]{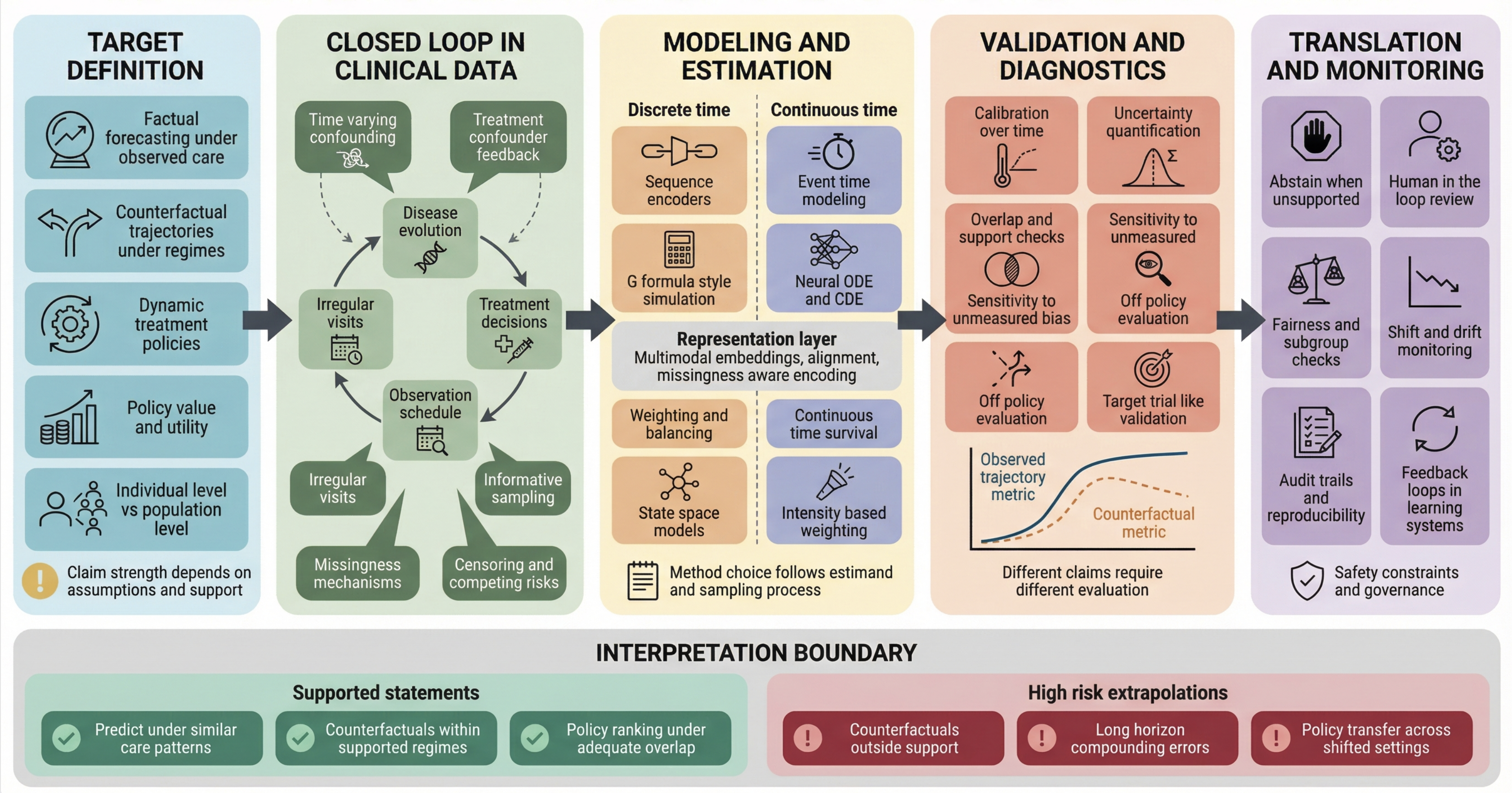}
 \caption{Discussion overview: assumptions, failure modes, and guardrails for intervention-aware disease trajectory modeling.}
 \label{fig:discussion}
\end{figure}

This chapter discusses intervention-aware trajectory modeling through a set of tightly linked themes that determine both scientific validity and clinical usefulness, rather than treating modeling choices as interchangeable engineering details. The discussion is structured around (i) targets and claims, clarifying whether the goal is forecasting under observed care, estimating counterfactual trajectories under specified regimes, or comparing dynamic policies, and what identifiability assumptions each claim entails; (ii) data-generating mechanisms, emphasizing the closed-loop coupling among disease evolution, treatment decisions, and the observation process, which introduces time-varying confounding, treatment–confounder feedback, and irregular or informative sampling; (iii) modeling and estimation, outlining how discrete-time and continuous-time formulations connect to these targets, and how representation learning, censoring, and missingness handling influence what is actually being learned; (iv) evaluation aligned with causal claims, distinguishing factual predictive metrics from counterfactual validation, overlap diagnostics, calibration over time, and off-policy evaluation; and (v) translation and safety, highlighting uncertainty-aware guardrails, limits of extrapolation beyond supported treatment histories, and monitoring for shift and drift after deployment in learning health systems. Figure~\ref{fig:discussion} provides a compact map of these themes and their dependencies.

\subsection{Reframing the field around intervention aware trajectory claims}
\label{sec:discussion-reframing}
Disease trajectory modeling is now a strong descriptive and predictive enterprise, yet it remains comparatively weak as an intervention aware science of longitudinal counterfactuals~\cite{intro_1,intro_3}.
A common disconnect is that many longitudinal models are evaluated with next event prediction or time updated risk objectives, while the scientific claim implicitly shifts toward counterfactual trajectories and decision improvement~\cite{intro_1,intro_3}.
The causal literature, by contrast, makes the target explicit through dynamic treatment regimes and time varying effects under sequential decision making, but these targets are not yet operationalized end to end in most modern trajectory pipelines~\cite{robins2000msm,2-17,murphy2003dtr}.
A concrete opportunity is to standardize intervention aware trajectory work around a small set of decision relevant estimands, such as trajectory distributions under explicit treatment rules, dynamic risk under specified policies, and policy value under realistic data support~\cite{robins2000msm,2-17,gottesman2019rlhc}.
This reframing is not a call for new terminology, it is a practical mechanism to prevent silent goal drift from forecasting to causal decision making~\cite{gottesman2019rlhc}.

\begin{table}[t]
\caption{Method selection matrix for trajectory modeling. Rows represent common research scenarios; columns provide actionable guidance. For causal applications (rows 4--10), the bottom row (validation requirements) applies universally.}
\centering
\footnotesize
\begingroup
\setlength{\tabcolsep}{3pt}
\renewcommand{\arraystretch}{1.08}
\begin{tabular}{@{}p{0.16\linewidth} p{0.18\linewidth} p{0.19\linewidth} p{0.24\linewidth} p{0.07\linewidth}@{}}
\hline
\textbf{Your situation} & \textbf{Recommended approach} & \textbf{Key methods} & \textbf{Critical considerations} & \textbf{Section} \\
\hline

Exploratory: Natural history, no treatment focus &
Multi-state, joint models &
Aalen--Johansen~\cite{aalen1978aj}; joint longitudinal model~\cite{wulfsohn1997joint_biometrics}; joint longitudinal--survival framework~\cite{rizopoulos2012joint_models_book} &
Interpretable, well-established, handles censoring &
\ref{subsec:trajectory-without-causal-semantics} \\

Prediction: Regular visits, dense data &
Deep EHR sequence models &
LSTM~\cite{hochreiter1997lstm}; RETAIN~\cite{choi2016retain}; BEHRT~\cite{li2020behrt}; Med-BERT~\cite{rasmy2021} &
Pretraining helps; check temporal shift &
\ref{subsec:trajectory-without-causal-semantics} \\

Prediction: Irregular sampling, sparse events &
Irregular-time models &
GRU-D~\cite{che2018missing}; Neural ODE~\cite{chen2018neuralode}; Neural CDE~\cite{kidger2020ncde}; Hawkes process~\cite{hawkes1971}; RMTPP~\cite{du2016rmttp}; Neural Hawkes~\cite{mei2017neural_hawkes} &
Model observation process if informative &
 \ref{rep}, ~\ref{subsec:trajectory-without-causal-semantics}\\

Causal: Short sequence (1--3 steps), moderate confounding &
Discrete-time g-methods or IPW &
RMSN~\cite{2-11}; G-Net~\cite{2-18}; MSM/IPW~\cite{robins2000msm} &
Check overlap at each step; weights can be unstable &
\ref{discrete}\\

Causal: Long sequence ($>$5 steps), complex history &
Balanced representations &
Causal Transformer~\cite{2-3}; CRN~\cite{2-2}; COSTAR~\cite{2-10} &
Balancing not always beneficial; needs large $N$ &
\ref{lss} \\

Causal: Continuous time, treatment timing matters &
Continuous-time counterfactuals &
TE-CDE~\cite{pmlr-v162-seedat22b}; CF-ODE~\cite{pmlr-v151-de-brouwer22a}; SCIP-Net~\cite{hess2024scipnet} &
Requires continuous-time assumptions; fragile extrapolation &
\ref{mctm} \\

Causal: Informative observation/measurement &
Observation-corrected models &
TESAR-CDE~\cite{pmlr-v202-vanderschueren23a} &
Must model visit intensity; adds complexity &
\ref{mctm} \\

Causal: Policy-level intervention &
Policy modeling + OPE &
Policy intervention model~\cite{pmlr-v202-hizli23a}; RL/OPE framework~\cite{pmlr-v68-raghu17a,gottesman2019rlhc} &
Extremely sensitive to overlap; conservative evaluation &
\ref{plmam} \\

Causal: Suspected hidden confounding &
Latent-factor / sensitivity methods &
Time Series Deconfounder~\cite{pmlr-v119-bica20a}; LipCDE~\cite{caoEWSMNL23}; continuous-time hidden-confounder extension~\cite{cao2023ct_hidden}; CDVAE~\cite{2-15} &
Identification fragile; sensitivity analysis mandatory &
\ref{fptdu}, ~\ref{mctm} \\

Causal: Poor overlap, safety-critical &
Uncertainty-aware models &
CF-ODE~\cite{pmlr-v151-de-brouwer22a}; Bayesian CDE~\cite{hess2023bncde}; distributional counterfactual model~\cite{2-4} &
Quantify epistemic uncertainty; restrict to overlap region &
\ref{fptdu}, ~\ref{mctm} \\

Any causal claim &
Validation requirements &
Overlap diagnostics~\cite{robins2000msm,2-17}; target trial emulation~\cite{hernan2016targettrial}; sensitivity analysis~\cite{2-17} &
Predictive metrics insufficient; need causal evaluation &
\ref{eval} \\

\hline
\end{tabular}
\endgroup
\label{tab:method-selection-matrix}
\end{table}

\subsection{Closing the loop in data generation}
\label{sec:discussion-closed-loop}
The dominant technical gap is that most models still do not close the full healthcare data-generation loop: latent disease progression influences clinician actions, actions alter outcomes, and both disease and actions alter what is measured and when~\cite{intro_1,2-17}. A related omission is that multimodal trajectory modeling is often acknowledged but not operationalized. In practice, clinically useful trajectories are assembled from heterogeneous streams, including high-frequency bedside physiology, lower-frequency imaging, free-text narratives, and sparse molecular markers. Reviews consistently show that multimodal setups improve clinical scope but remain under-specified in temporal and causal terms~\cite{kline2022,wuuhar,huang2020fusion_ehr_imaging}.

Current multimodal integration can be organized into three design families. First, shared-state early fusion maps all modalities into a common latent patient state before temporal prediction; this is simple but can blur modality-specific uncertainty. Second, modality-specific encoders with cross-modal fusion preserves structure (e.g., CNN/ViT for imaging, transformer/RNN for text and EHR sequences, graph encoders for omics pathways) and then fuses representations with attention or gating~\cite{rajkomar2018,zhang2020structured_unstructured,cheng2017ccrcc_integrative,cheerla2019pancancer}. Third, late-fusion ensembles combine modality-level predictions and are often more robust to missing modalities, though they may underuse cross-modal interactions~\cite{kline2022,huang2020fusion_ehr_imaging}.

Coordinating asynchronous sampling rates is the core temporal challenge. A practical formulation is a continuous latent state with event-triggered modality updates,
\begin{equation}
\frac{d z(t)}{dt}=f_\theta\big(z(t),t\big),
\qquad
z(t^+)=z(t^-)+\sum_{m\in\mathcal M} I_m(t)\,g_m\!\big(x_m(t)\big),
\end{equation}
where $I_m(t)$ indicates whether modality $m$ is observed at time $t$. For example, daily vital signs can provide dense updates to $z(t)$, weekly imaging can induce larger structural corrections, and quarterly biomarker/omics panels can recalibrate slower disease axes. Continuous-time encoders and observation-process-aware methods are particularly suitable for this regime because they decouple latent dynamics from irregular observation times~\cite{rubanova2019latentode,kidger2020ncde,pmlr-v202-vanderschueren23a,pmlr-v162-seedat22b}.

Key limitations remain. First, multimodal missingness is typically informative rather than random: imaging is often ordered after deterioration, and omics is often collected in selected subcohorts, which can induce selection bias if not modeled explicitly. Second, modality imbalance can cause representation domination by high-frequency channels (for example vitals), suppressing weak but clinically meaningful slow signals (for example interval imaging or quarterly biomarkers). Third, external transport is fragile because scanner protocols, note styles, assay platforms, and coding practices shift across institutions~\cite{kline2022,huang2020fusion_ehr_imaging,wiens2019donoHarm,feuerriegel2024natmed}.

A high-value next step is \emph{causally structured multimodal trajectory modeling}: explicitly parameterizing how each modality enters disease-state estimation, treatment policy estimation, and observation intensity. This enables modality-level sensitivity analysis, overlap diagnostics by modality, and uncertainty decomposition that distinguishes data sparsity from model extrapolation. Another promising direction is adaptive sensing policies that decide \emph{when} to request expensive modalities (for example imaging or omics) based on expected decision value, thereby linking trajectory modeling to resource-aware clinical workflows. Crucially, claims should be validated with target-trial-aligned protocols and, where feasible, prospective randomized or quasi-experimental components, so multimodal gains are judged by decision impact rather than headline AUROC alone~\cite{hernan2016targettrial,gottesman2019rlhc,pmlr-v151-de-brouwer22a}.

\subsection{Evaluation that matches the claim, trajectories, policies, and uncertainty}
\label{eval}
Evaluation remains the binding constraint because trajectory-level and policy-level claims cannot be justified by retrospective discrimination metrics alone~\cite{intro_1,intro_3,gottesman2019rlhc}. For factual prediction, a Nature-standard evaluation should report \emph{time-indexed} performance at clinically meaningful decision times, rather than a single pooled AUROC. A common dynamic discrimination target is
\begin{equation}
\mathrm{AUC}(t,h)=P\!\left(\hat r_i(t,h)>\hat r_j(t,h)\mid T_i\in(t,t+h],\,T_j>t+h\right),
\end{equation}
where $t$ is the landmark time, $h$ is the prediction horizon, and $T$ is event time~\cite{heagerty2000roc,heagerty2005roc}. Reporting $\mathrm{AUC}(t,h)$ as a trajectory reveals whether discrimination degrades exactly when decisions are most consequential.

Discrimination must be paired with probabilistic accuracy and calibration over time. With right-censoring, a weighted Brier trajectory can be written as
\begin{equation}
\mathrm{BS}(t,h)=\frac{1}{N}\sum_{i=1}^{N}w_i(t,h)\left(\mathbf{1}\{T_i\le t+h\}-\hat p_i(t,h)\right)^2,
\end{equation}
which quantifies squared probability error at each decision time and horizon~\cite{graf1999brier,gerds2006brier}. Calibration should likewise be treated as a time-varying function,
\begin{equation}
m_{t,h}(p)=P\!\left(T\le t+h\mid T>t,\hat p(t,h)=p\right),
\end{equation}
estimated with dynamic calibration curves plus summary indices (e.g., slope/intercept or integrated calibration error), because well-ranked but miscalibrated risk trajectories can still induce unsafe treatment choices~\cite{vancalster2019calibration,austin2020calibration}.

Counterfactual validation requires designs that introduce external variation beyond model assumptions. Three practical routes are complementary rather than competing. First, a randomized subset strategy can embed prospective randomization or randomized encouragement among clinically eligible/overlap patients, yielding a local benchmark for treatment-effect direction, magnitude, and policy ranking~\cite{relton2010cmrct,hernan2016targettrial}. Second, instrumental-variable validation can exploit plausibly exogenous treatment variation $Z_t$,
\begin{equation}
A_t=\alpha_0+\alpha_1 Z_t+\alpha_2 H_t+\varepsilon_t,\qquad
Y_{t+1}=\beta_0+\beta_1\hat A_t+\beta_2 H_t+\eta_t,
\end{equation}
to test whether model-implied effects are consistent with IV-identified effects under relevance, exclusion, and independence assumptions~\cite{angrist1996iv,baiocchi2014iv}. Third, natural experiments (for example protocol rollouts or exogenous service shocks) support quasi-experimental triangulation through interrupted-time or difference-in-differences logic, reducing dependence on one modeling pipeline~\cite{craig2012natural,hernan2016targettrial}.

An actionable example is \emph{sepsis trajectory model evaluation by stage}. Stage 1 (cohort and time-zero definition): enforce Sepsis-3 compatible entry criteria, define intervention windows (e.g., fluids/vasopressors/antibiotics) and avoid immortal-time leakage~\cite{singer2016sepsis3,evans2021sepsis}. Stage 2 (factual trajectory fidelity): report $\mathrm{AUC}(t,h)$, $\mathrm{BS}(t,h)$, and dynamic calibration across early shock phases and clinically relevant subgroups. Stage 3 (counterfactual identifiability): audit overlap, propensity extremity, weight truncation sensitivity, and hidden-confounding robustness before interpreting effect heterogeneity~\cite{robins2000msm,2-17}. Stage 4 (policy-level validity): evaluate candidate policies with uncertainty intervals, stress-test off-policy estimators, and check whether policy gains persist under alternative estimators and support restrictions~\cite{komorowski2018aiclinician,gottesman2019rlhc}. Stage 5 (deployment safety): monitor post-deployment drift in calibration and policy value, and require abstention/escalation when support deteriorates, so evaluation remains a continuous safety process rather than a one-time benchmark~\cite{wiens2019donoHarm,feuerriegel2024natmed}.

Under this framework, evaluation is not a terminal scoreboard but a causal validity pipeline: each stage asks whether a stronger claim (forecasting $\rightarrow$ counterfactuals $\rightarrow$ policy) is empirically supported by design, diagnostics, and uncertainty quantification.

\subsection{Interpretability as a causal constraint}
\label{sec:discussion-interpretability}
Interpretability is part of the safety case for acting on model outputs over time, not an optional visualization step~\cite{intro_1,gottesman2019rlhc}.
Classical causal methods provide transparent estimands and assumptions, whereas deep temporal models offer flexible representations but can obscure failure modes such as extrapolation under poor overlap and latent entanglement of disease and care pathways~\cite{robins2000msm,2-17,chen2018neuralode,kidger2020ncde}.
A promising opportunity is structured hybridization, imposing clinically anchored state structure or constrained transition modules within expressive deep models, so that counterfactual trajectories remain auditable while still leveraging high dimensional EHR covariates~\cite{intro_1,robins2000msm,2-17}.
The practical payoff is twofold, clinicians gain interpretable trajectory summaries, and researchers gain levers for sensitivity analysis that are difficult to conduct in monolithic end to end black boxes~\cite{2-17,gottesman2019rlhc}. 

\subsection{Synthesis of evidence across the research questions}
\label{sec:rq-synthesis}
To close the argument, we answer the four research questions as a connected progression: target definition (RQ1) $\rightarrow$ identification (RQ2) $\rightarrow$ estimation (RQ3) $\rightarrow$ decision-grade translation (RQ4).

\begin{enumerate}
  \item \textbf{Addressing RQ1 (scientific target).} Section~\ref{sec:conceptual-review-landscape} and Subsections~\ref{3.1}--\ref{3.2} reframe trajectories as dynamic clinical objects rather than static risk snapshots. Subsection~\ref{section:Causal} then formalizes this object using potential-outcome trajectories under dynamic treatment regimes, making treatment and observation mechanisms part of the estimand rather than auxiliary covariates. Discussion Subsection~\ref{sec:discussion-reframing} sharpens the claim hierarchy (forecasting, counterfactual, policy) and identifies claim inflation as a major source of error. The resulting answer is that the primary scientific target is a joint intervention-aware trajectory estimand, not a single discriminative score.

  \item \textbf{Addressing RQ2 (identifiability conditions).} Subsection~\ref{section:Causal} provides the identifiability scaffold through sequential exchangeability, positivity, and consistency. Sections~\ref{sec:discrete-intervention} and~\ref{sec:continuous-policy} operationalize these conditions in discrete and continuous time through weighting/balancing strategies, continuous-time adjustment, and observation-process-aware modeling. Discussion Subsection~\ref{sec:discussion-closed-loop} deepens this by linking assumptions to realistic violations, including informative sampling, modality-dependent missingness, and transport instability. The synthesis is assumption-first: trajectory claims are credible only within explicitly diagnosed support and data-generation regimes.

  \item \textbf{Addressing RQ3 (estimation strategy).} Methodological evidence is organized as estimand-aligned families rather than as an architecture catalog. Section~\ref{sec:discrete-intervention} develops discrete-time intervention-aware sequence learning for dynamic regimes and long-horizon counterfactuals, while Section~\ref{sec:continuous-policy} extends to irregular-time dynamics and policy-level intervention semantics. Section~\ref{subsec:trajectory-without-causal-semantics} serves as a boundary condition by showing where high-performing trajectory predictors remain prognostic yet non-interventional. Together with Discussion Subsections~\ref{sec:discussion-reframing} and~\ref{sec:discussion-closed-loop}, this establishes that model choice is primarily constrained by estimand clarity, support, and observation-process structure, with representational expressiveness as a secondary criterion.

  \item \textbf{Addressing RQ4 (decision-grade evidence and translation).} Subsection~\ref{eval} specifies a decision-grade evidence ladder from time-indexed factual fidelity to counterfactual credibility and policy-level value, with explicit requirements for dynamic calibration, overlap diagnostics, uncertainty quantification, and off-policy robustness. Subsection~\ref{sec:discussion-interpretability} adds interpretability as a causal safety constraint rather than a post hoc visualization preference. Subsection~\ref{sec:discussion-closed-loop} and Section~\ref{sec:discussion} extend evaluation into deployment through drift monitoring, support deterioration checks, and abstention triggers. The resulting translation criterion is strict: deployment is justified only when intended claim strength, empirical support, and safety governance remain aligned over time.
\end{enumerate}

\section{Conclusion}
This review advocates a principled shift in clinical ML, moving from snapshot-based risk scoring toward dynamic, intervention-aware disease trajectory modeling. Static predictors can support screening and stratification, yet they are often misaligned with the questions that drive clinical decisions. Clinicians need to understand how a patient’s course may evolve over time, how treatment choices shape that evolution, and what outcomes might be expected under alternative regimens or policies. In longitudinal care, disease progression, treatment decisions, and measurement schedules interact in a closed loop. When this coupling is ignored and interventions are treated as ordinary features, models can look accurate for forecasting under usual care while becoming unreliable for counterfactual reasoning.

A central takeaway is that trustworthy deployment depends less on model expressiveness than on alignment among learning targets, data-generating mechanisms, and evaluation. Time-varying confounding, treatment–confounder feedback, irregular and informative observation, and distribution shift across sites and care pathways can all distort conclusions if treated as secondary nuisances. The most useful frameworks therefore make assumptions explicit, define estimands precisely, and select modeling strategies that respect the temporal semantics of interventions, whether represented in discrete time, continuous time, or event-driven formulations.

Future progress requires integrating three components into a coherent pipeline. First, closed-loop modeling should jointly represent latent disease evolution, treatment assignment, and the observation process rather than treating them as independent. Second, uncertainty and safety guardrails should become first-class outputs. Models should quantify epistemic uncertainty, diagnose overlap and positivity failures, and abstain when counterfactual support is weak. Third, evaluation must match the causal claim. Validation protocols should go beyond predictive metrics on observed trajectories, use designs that approximate target trials where appropriate, assess robustness to plausible unmeasured biases, and evaluate policy performance with careful attention to off-policy error.

In summary, intervention-aware trajectory modeling provides a unifying perspective for clinical AI by connecting multimodal longitudinal data to actionable forecasting, counterfactual analysis, and policy evaluation under transparent assumptions. Success is best measured not only by higher predictive accuracy, but by reliable decision support that characterizes how outcomes may change under plausible clinical actions and clearly communicates when evidence is insufficient to favor one action over another.

\bibliographystyle{plain}
\bibliography{references}

\end{document}